\RequirePackage{fix-cm}
\documentclass[twocolumn]{svjour3}          
\smartqed  

\usepackage{times}
\usepackage{epsfig}
\usepackage{graphicx}
\usepackage{amsmath}
\usepackage{amssymb}
\usepackage[boxed]{algorithm2e}
\usepackage{subfigure}
\usepackage{multirow}
\usepackage{color}
\usepackage{hyperref}

\newcommand{\yj}{\textcolor{black}}
\begin{document}
\sloppy

\title{Predicting Important Objects for Egocentric Video Summarization
}


\author{Yong~Jae~Lee         \and
        Kristen~Grauman 
}


\institute{Yong~Jae~Lee \at
              Department of Computer Science, \\
              University of California, Davis. \\
              \email{yjlee@cs.ucdavis.edu}           
           \and
           Kristen~Grauman \at
              Department of Computer Science, \\
              University of Texas at Austin.\\
              \email{grauman@cs.utexas.edu}
}

\date{Received: date / Accepted: date}

\maketitle

\begin{abstract}
We present a video summarization approach for egocentric or ``wearable'' camera data.  Given hours of video, the proposed method produces a compact storyboard summary of the camera wearer's day.  In contrast to traditional keyframe selection techniques, the resulting summary focuses on the most important objects and people with which the camera wearer interacts.  To accomplish this, we develop region cues indicative of high-level saliency in egocentric video---such as the nearness to hands, gaze, and frequency of occurrence---and learn a regressor to predict the relative importance of any new region based on these cues.  Using these predictions and a simple form of temporal event detection, our method selects frames for the storyboard that reflect the key object-driven happenings.  We adjust the compactness of the final summary given either an importance selection criterion or a length budget; for the latter, we design an efficient dynamic programming solution that accounts for importance, visual uniqueness, and temporal displacement.  Critically, the approach is neither camera-wearer-specific nor object-specific; that means the learned importance metric need not be trained for a given user or context, and it can predict the importance of objects and people that have never been seen previously.  \yj{Our results on two egocentric video datasets} show the method's promise relative to existing techniques for saliency and summarization.

\end{abstract}

\section{Introduction}
\label{intro}
{T}{he} goal of video summarization is to produce a compact visual summary that encapsulates the key components of a video.  Its main value is in turning hours of video into a short summary that can be interpreted by a human viewer in a matter of seconds.  Automatic video summarization methods would be useful for a number of practical applications, such as analyzing surveillance data, video browsing, action recognition, or creating a visual diary of one's personal lifelog video.

Existing methods extract keyframes~\cite{wolf-1996,zhang-pr1997,goldman-2006,Liu-2002}, create montages of still images~\cite{aner-2002,caspi-2006}, or generate compact dynamic summaries~\cite{ravacha-2006,pritch-2007}.  Despite promising results, they assume a static background or rely on low-level appearance and motion cues to select what will go into the final summary.  However, in many interesting settings, such as egocentric videos, YouTube style videos, or feature films, the background is moving and changing.  More critically, a system that lacks high-level information on \emph{which objects matter} may produce a summary that consists of irrelevant frames or regions.  In other words, existing methods are indifferent to the impact that each object has on generating the ``story'' of the video.

\begin{figure}[t]
\centering
\hspace*{-0.1in}
\includegraphics[width=8.5cm]{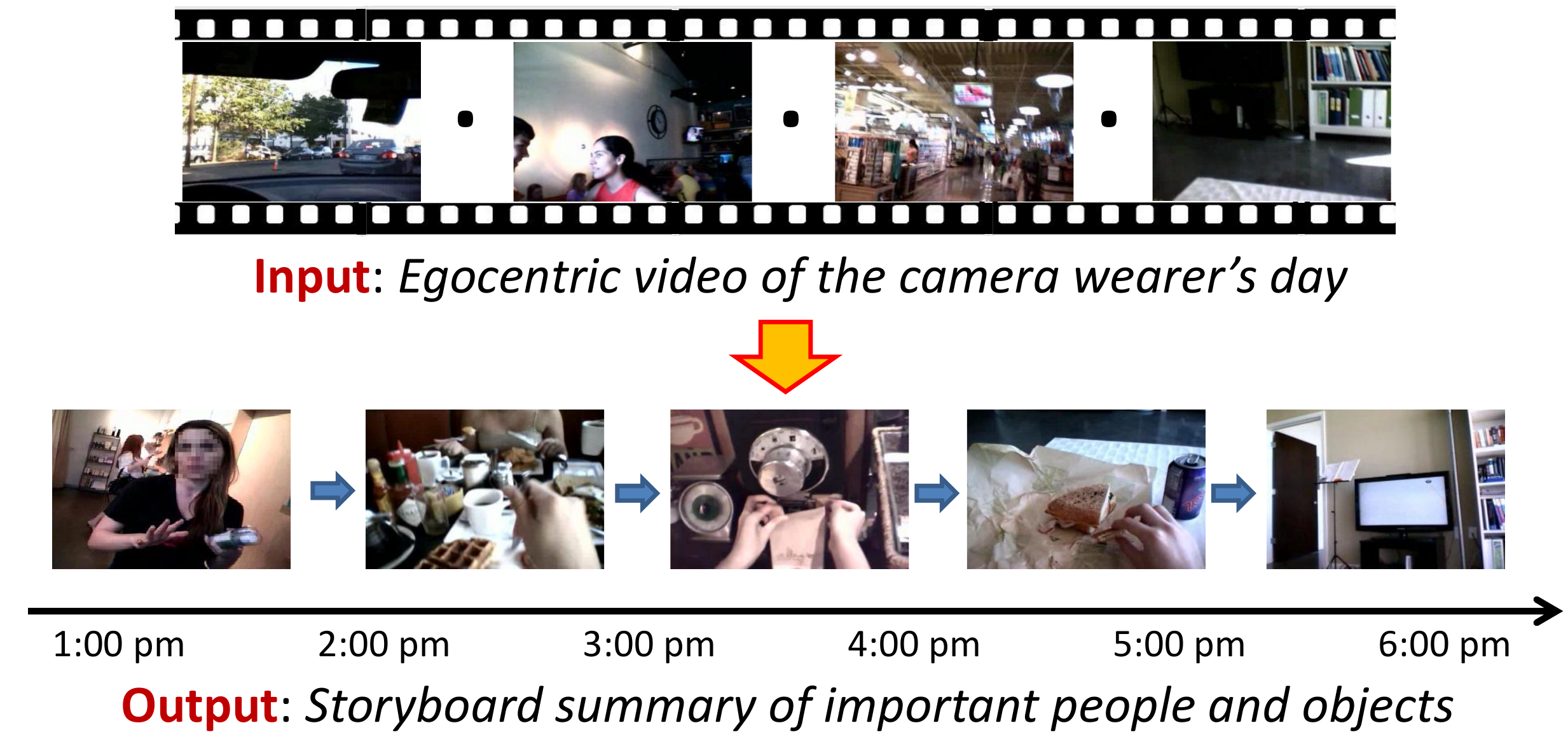}
\caption{Given an unannotated egocentric video, our method produces a compact storyboard visual summary that focuses on the key people and objects.}
\label{fig:concept}
\end{figure}

In this work, we are interested in creating object-driven summaries for videos captured from a wearable camera.  An egocentric video offers a first-person view of the world that cannot be captured from environmental cameras.  For example, we can often see the camera wearer's hands, or find the object of interest centered in the frame.  Essentially, a wearable camera focuses on the user's activities, social interactions, and interests.  We aim to exploit these properties for egocentric video summarization.

Good summaries for egocentric data would have wide potential uses.  Not only would recreational users (including ``life-loggers") find it useful as a video diary, but there are also high-impact applications in law enforcement, elder and child care, and mental health.  For example, the summaries could facilitate police officers in reviewing important evidence, suspects, and witnesses, or aid patients with memory problems to remember specific events, objects, and people\yj{~\cite{hodges-memory2011,lee-2007}}.  Furthermore, the egocentric view translates naturally to robotics applications---suggesting, for example, that a robot could summarize what it encounters while navigating unexplored territory, for later human viewing.

Motivated by these problems, we propose an approach that learns category-independent \emph{importance} cues designed explicitly to target the \emph{key objects and people} in the video.  The main idea is to leverage novel egocentric and high-level saliency features to train a model that can predict important regions in the video, and then to produce a concise visual summary that is driven by those regions (see Fig.~\ref{fig:concept}).  By learning to predict important regions, we can focus the visual summary on the main people and objects, and ignore irrelevant or redundant information.

Our method works as follows.  We first train a regression model from labeled training videos that scores any region's likelihood of belonging to an important person or object.  For the input variables, we develop a set of high-level cues to capture egocentric importance, such as frequency, proximity to the camera wearer's hand, and object-like appearance and motion, as well as a set of low-level cues to capture region properties such as size, width, and height.  The target variable is the overlap with ground-truth important regions, i.e., the \emph{importance score}.  Given a novel video, we use the model to predict important regions for each frame.  We then partition the video into unique temporal \emph{events}, by clustering scenes that have similar color distributions and are close in time.  For each event, we isolate unique representative instances of each important person or object.  Finally, we produce a storyboard visual summary that displays the most important objects and people across all events in the camera wearer's day.

We propose two ways to adjust the compactness of the summary, based on either a target importance criterion or a target summary length.  For the latter, we design an energy function that accounts for the importance of the selected frames, their visual dissimilarities, and their temporal displacements, and can be efficiently optimized using dynamic programming.

We emphasize that we do not aim to predict importance for any specific category (e.g., cars).  Instead, we learn a general model that can predict the importance of any object instance, irrespective of its category.  This category-independence avoids the need to train importance predictors specific to a given camera wearer, and allows the system to recognize as important something it has never seen before.  In addition, it means that objects from the same category can be predicted to be (un)important depending on their role in the story of the video.  For example, if the camera wearer has lunch with his friend Jill, she would be considered important, whereas people in the same restaurant sitting around them could be unimportant.  Then, if they later attend a party but chat with different friends, Jill may no longer be considered important in that context.

Our main contribution is an egocentric video summarization approach that is driven by predicted important people and objects.  Towards this goal, we develop two primary technical ideas.  In the first, we develop a learning approach to estimate region importance using novel cues designed specifically for the egocentric video setting.  In the second, we devise an efficient keyframe selection strategy that captures the most important objects and people, subject to meeting a budget for the desired length of the output storyboard.

We apply our method to challenging real-world videos captured by users in uncontrolled environments, and process a total of \yj{27 hours of video}---significantly more data than previous work in egocentric analysis.  Evaluating the predicted importance estimates and summaries, we find our approach outperforms state-of-the-art high-level and low-level saliency measures for this task, and produces significantly more informative summaries than traditional methods.

This article expands upon our previous conference paper~\cite{egocentric} in terms of the method design, experiments, and presentation.  In Sections~\ref{subsec:dp} and~\ref{subsec:lengthBudgetAccuracy}, we introduce and analyze a novel budgeted frame selection approach that efficiently produces fixed-length summaries.  \yj{In Section~\ref{sec:results}, we add new comparisons to multiple existing video summarization methods, analyze object prominence in the summaries, conduct new user studies with over 25 users to systematically gauge the summaries' quality, and produce new results on the Activities of Daily Living dataset~\cite{deva-egocentric-cvpr2012}.} Finally, throughout we provide more detailed algorithm explanations (including Figures~\ref{fig:interface},~\ref{fig:features}, and~\ref{fig:regionGrouping}).

\section{Related Work}
\label{sec:relatedwork}

\textbf{Video summarization:  }Static keyframe methods compute motion stability from optical flow~\cite{wolf-1996} or global scene color/texture differences~\cite{zhang-pr1997,Liu-2002,icme2009} to select the frames that go into the summary.  The low-level approach means that irrelevant frames can often be selected, which is particularly problematic for our application of summarizing hours of continuous egocentric video that contain lots of irrelevant data.  By generating object-driven summaries, we aim to move beyond such low-level cues.

Video summarization can also take the form of a single montage of still images.  Existing methods take a background reference frame and project in foreground regions~\cite{aner-2002}, or sequentially display automatically selected key-poses~\cite{caspi-2006}.  An interactive approach~\cite{goldman-2006} takes user-selected frames and key points, and generates a storyboard that conveys the trajectory of an object.  These approaches generally assume short clips with few objects, or a human-in-the-loop to guide the summarization process.  In contrast, we aim to summarize a camera wearer's day containing hours of continuous video with hundreds of objects, with no human intervention.

Compact dynamic summaries simultaneously show several spatially non-overlapping actions from different times of the video~\cite{ravacha-2006,pritch-2007}.  While that framework aims to focus on foreground objects, it assumes a static camera and is therefore inapplicable to egocentric video.  A re-targeting approach aims to simultaneously preserve an original video's content while reducing artifacts~\cite{simakov-cvpr2008}, but unlike our approach, does not attempt to characterize the varying degrees of object importance.  In a semi-automatic method~\cite{liu-pami2009}, irrelevant video frames are removed by detecting the main object of interest given a few user-annotated training frames.  In contrast, our approach \emph{automatically} discovers multiple important objects.

\textbf{Saliency detection:  }Early saliency detectors rely on bottom-up image cues (e.g.,~\cite{itti-1998,gao-nips2007}).  More recent work tries to learn high-level saliency measures using various Gestalt cues, whether for static images~\cite{liu-cvpr07,objectness,carreira-mincut,endres-eccv2010} or video~\cite{keysegments}.  Whereas typically such metrics aim to prime a visual search process, we are interested in high-level saliency for the sake of isolating those things worth summarizing.  Researchers have also explored ranking object importance in static images, learning what people mention first from human-annotated tags~\cite{spain-eccv2008,hwang-bmcv2010}.  In contrast, we learn the importance of objects in terms of their role in a long-term video's story.  Relative to any of the above, we introduce novel saliency features amenable to the egocentric video setting.

\textbf{Egocentric visual data analysis:  }Vision researchers have recently returned to exploring egocentric visual analysis, prompted in part by increasingly portable wearable cameras.  Early work with wearable cameras partition visual and audio data into events~\cite{clarkson-icassp1999}, or uses supervised learning for specialized tasks like sign language recognition~\cite{starner-pami1998} or location recognition within a building~\cite{starner-iswc1998}.  \yj{Methods in ubiquitous computing use manual intervention~\cite{mann-1998} or external non-visual sensors~\cite{healey-1998,hodges-2006} (e.g., skin conductivity or audio) to trigger snapshots from a wearable camera.  Others use brain waves~\cite{ng-icme02}, k-means clustering with temporal constraints~\cite{Lin06a.:structuring}, or face detection~\cite{doherty-2008} to segment egocentric videos.  Recent methods explore activity recognition~\cite{spriggs-wkshp,fathi-iccv2011,deva-egocentric-cvpr2012,ryoo-cvpr2013}, handled object recognition~\cite{ren-cvpr2010}, novelty detection~\cite{aghazadeh-cvpr2011}, hand detection~\cite{li-cvpr2013}, gaze prediction~\cite{li-iccv2013}, social interaction analysis~\cite{fathi-cvpr2012}, or activity discovery for non-visual sensory data~\cite{schiele-ubicomp08}.}  Unsupervised algorithms are developed to discover scenes~\cite{jojic-2010} and actions~\cite{kitani-2011}, or select keyframes~\cite{doherty-civr2008}, based on low-level visual features extracted from egocentric data.  In contrast to all these methods, we aim to build a visual summary, and model high-level importance of the objects present.

\yj{To our knowledge, we are the first to explore visual summarization of egocentric video by predicting important objects.  Recent work~\cite{lu-cvpr2013} builds on our approach and uses our importance predictions as a cue to generate story-driven egocentric video summarizations.}

\section{Approach}
\label{sec:approach}

Our goal is to create a storyboard summary of a person's day that is driven by the important people and objects.  The video is captured using a wearable camera that continuously records what the user sees.  We define \emph{importance} in the scope of egocentric video: important things are those with which the camera wearer has significant interaction.  This is reasonable for the egocentric setting, since the camera wearer is likely to engage in social activities with cliques of people (e.g., friends, co-workers) that involve interactions with specific objects (e.g., food, computer).  The camera wearer will typically find these people and objects to be memorable, as we confirm in our user studies in Section~\ref{subsec:userstudies}.

There are four main steps to our approach: (1) using novel egocentric saliency cues to train a category-independent regression model that predicts how likely  it is that an image region belongs to an important person or object; (2) partitioning the video into temporal events.  For each event, (3) scoring each region's importance using the regressor; and (4) selecting representative key-frames for the storyboard that encapsulate the predicted important people and objects, either using a user-specified importance criterion or a length budget.

We first describe how we collect the video data and ground-truth annotations needed to train our model.  We then describe each of the main steps in turn.

\begin{figure*}[t!]
\centering
\includegraphics[width=17cm]{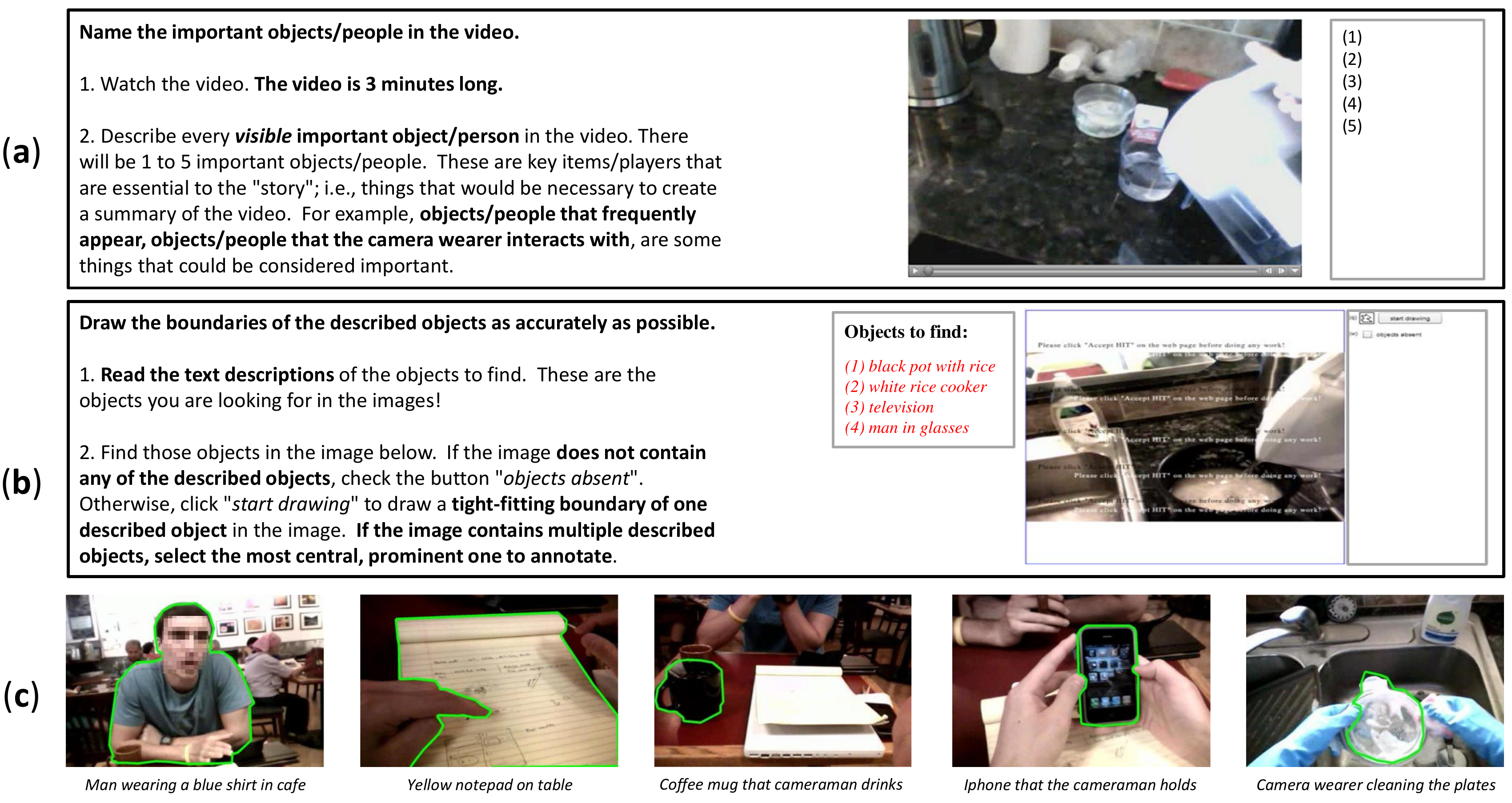}
\caption{Our Mechanical Turk interfaces for important person/object (a) text description and (b) annotation, and (c) example annotations that we obtained.  The important people and objects are annotated.}
\label{fig:interface}
\end{figure*}

\subsection{Egocentric video data collection}
\label{subsec:dataCollection}

We use the Looxcie wearable camera, which captures video at 15 fps at 320 x 480 resolution.  It is worn around the ear and looks out at the world at roughly eye-level.  We collected 10 videos from four subjects, each three to five hours in length (the maximum battery life), for a total of 37 hours of video.  We call this the UT Egocentric (UT Ego) dataset.  Our data is publicly available.\footnote{\scriptsize{\texttt{http://vision.cs.utexas.edu/projects/egocentric/}}\\Due to privacy issues, we are only able to share 4 of the 10 videos (one from each subject), for a total of 17 hours of video.  They correspond to the test videos that we evaluate on in Sec.~\ref{sec:results}.}

Four subjects wore the camera for us: one undergraduate student, two grad students, and one office worker, ranging in age from early to late 20s and both genders.  The different backgrounds of the subjects ensure diversity in the data---not everyone's day is the same---and is critical for validating the category-independence of our approach.  We asked the subjects to record their natural daily activities, and explicitly instructed them not to stage anything for this purpose.  The videos capture a variety of activities such as eating, shopping, attending a lecture, driving, cooking, and working on a computer.

\subsection{Annotating important regions in video}
\label{subsec:mturk}

To train the importance predictor, we first need ground-truth training examples.  In general, determining whether an object is important or not can be highly subjective.  Fortunately, an egocentric video provides many constraints that are suggestive of an object's importance.  For example, one can observe the camera wearer's hands, and an object of interest may often be centered in the frame.

In order to learn meaningful egocentric properties without overfitting to any particular category, we crowd-source annotations using Amazon's Mechanical Turk (MTurk).  For egocentric videos, an object's degree of importance will depend on what the camera wearer is doing before, while, and after the object or person appears.  In other words, the object must be seen in the context of the camera wearer's activity to properly gauge its importance.

We carefully design two annotation tasks to capture this aspect.  In the first task, we ask workers to watch a three minute accelerated video (equivalent to 10 minutes of original video) and to describe in text what they perceive to be essential people or objects necessary to create a summary of the video.  In the second task, we display uniformly sampled frames from the video and their corresponding text descriptions \emph{obtained from the first task}, and ask workers to draw polygons around any described person or object.  If none of the described objects are present in a frame, the annotator is given the option to skip it.  See Fig.~\ref{fig:interface} for the two interfaces and example annotations. 

We found this two-step process more effective than a single task in which the same worker both watches the video and then annotates the regions s/he deems important, likely due to the time required to complete both tasks.  Critically, the two-step process also helps us avoid bias: a single annotator asked to complete both tasks at once may be biased to pick easier things to annotate rather than those s/he finds to be most important.  Our setup makes it easy for the first worker to freely describe the objects without bias, since s/he only has to enter text.  We found the resulting annotations quite consistent, and only manually pruned those where the region outlined did not agree with the first worker's description.  For a 3-5 hour training video, we obtain roughly 35 text descriptions and 700 object segmentations.

\subsection{Learning egocentric region importance}
\label{subsec:learning}

We now discuss the procedure to train a general purpose category-independent model that will predict important regions in any egocentric video, independent of the camera wearer.  Given a video, we first generate candidate regions for each frame using a min-cut method~\cite{carreira-mincut}, which tends to avoid oversegmenting objects.  We represent objects at the frame-level, since our uncontrolled setting usually prohibits reliable space-time object segmentation due to frequent and rapid head movements by the camera wearer.  We generate roughly 800 regions per frame.

For each region, we compute a set of candidate features that could be useful to describe its importance.  Since the video is captured by an active participant, we specifically want to exploit egocentric properties such as whether the object/person is interacting with the camera wearer, whether it is the focus of the wearer's gaze, and whether it frequently appears.  In addition, we aim to capture high-level saliency cues---such as an object's motion and appearance, or the likelihood of being a human face---and generic region properties shared across categories, such as size or location.  We describe the proposed features in detail next.

\subsubsection{Feature definitions}
\label{subsubsec:features}

\begin{figure}[t]
\centering
\includegraphics[width=8.2cm]{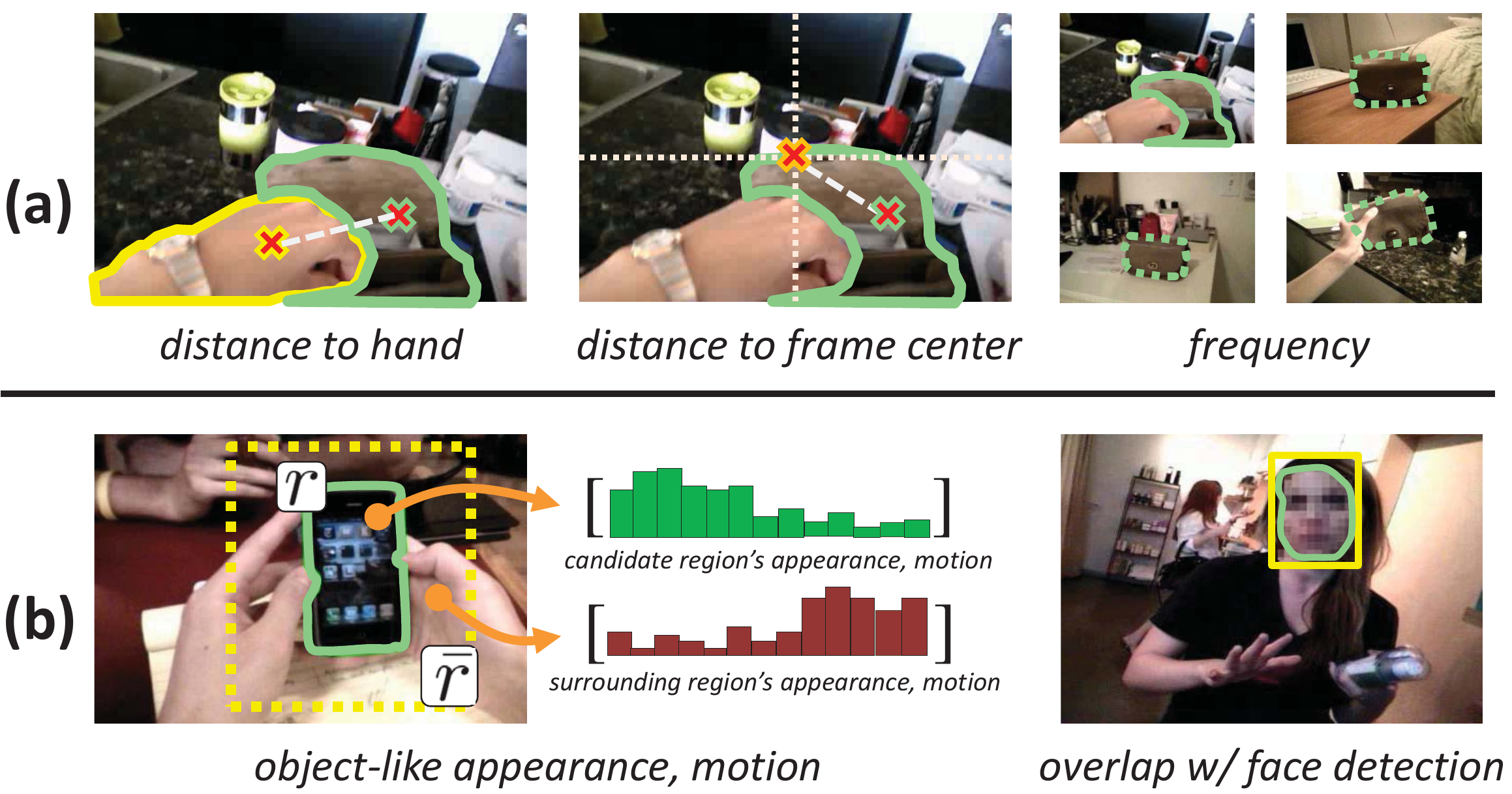}
\caption{Illustration of our (a) egocentric features and (b) object features.}%
\label{fig:features}
\end{figure}

\textbf{\emph{Egocentric features}:  }Fig.~\ref{fig:features} (a) illustrates the three proposed egocentric features.  To model \textbf{interaction}, we compute the Euclidean distance of the region's centroid to the closest detected hand in the frame.  Given a frame in the test video, we first classify each pixel as (non-)skin using color likelihoods and a Naive Bayes classifier~\cite{skindetection} trained with ground-truth hand annotations on disjoint data.  We then classify any superpixel (computed using~\cite{pedro-superpixels}) as hand if more than 25\% of its pixels are skin.  While simple, we find this hand detector is sufficient for our application.  More sophisticated methods (e.g.,~\cite{turk-fg2004}) would certainly be possible as well.

To model \textbf{gaze}, we compute the Euclidean distance of the region's centroid to the frame center.  Since the camera moves with the wearer's head, this is a coarse estimate of how likely the region is being focused upon.

To model \textbf{frequency}, we record the number of times an object instance is detected within a short temporal segment of the video.  We create two frequency features: one based on  matching regions, the other based on matching points.  For the first, we compute the color dissimilarity between a region $r$ and each region $r_n$ in its surrounding frames, and accumulate the total number of positive matches:
\begin{equation}
c_{region}(r) = \sum_{f \in \mathcal{W}} [\Bigl(\min_n \chi^2(r, r_n^f)\Bigr) \le \theta_r],
\label{eqn:regionFrequency}
\end{equation}
where $f$ indexes the set of frames $\mathcal{W}$ surrounding region $r$'s frame, $\chi^2(r, r_n)$ is the $\chi^2$-distance between color histograms of $r$ and $r_n$, $\theta_r$ is the distance threshold to determine a positive match, and $[\cdot]$ denotes the indicator function.  The value of $c_{region}$ will be high/low when $r$ produces many/few matches (i.e., is frequent/infrequent).

The second frequency feature is computed by matching Difference of Gaussian SIFT interest points.  For a detected point $p$ in region $r$, we match it to all detected points in each frame $f \in \mathcal{W}$, and count as positive those that pass the ratio test~\cite{lowe}.  We repeat this process for each point in region $r$, and record their average number of positive matches:
\begin{equation}
c_{point}(r) = \frac{1}{P} \sum_{i=1}^P \sum_{f \in \mathcal{W}} \left[\frac{d(p_i,p_{1^*}^f)}{d(p_i,p_{2^*}^f)} \le \theta_p \right],
\label{eqn:pointFrequency}
\end{equation}
where $i$ indexes all detected points in region $r$, $d(p_i,p_{1^*}^f)$ and $d(p_i,p_{2^*}^f)$ measure the Euclidean distance between $p_i$ and its best matching point $p_{1^*}^f$ and second best matching point $p_{2^*}^f$ in frame $f$, respectively, and $\theta_p$ is Lowe's ratio test threshold for non-ambiguous matches~\cite{lowe}.  The value of $c_{point}$ will be high/low when the SIFT points in $r$ produce many/few matches.  For both frequency features, we set $\mathcal{W}$ to span a 10 minute temporal window.

\textbf{\emph{Object features}:  }In addition to the egocentric-specific features, we include three high-level (i.e., object-based) saliency cues (see Fig.~\ref{fig:features} (b)).  To model \textbf{object-like appearance}, we use the learned region ranking function of~\cite{carreira-mincut}.  It reflects Gestalt cues indicative of \emph{any} object, such as the sum of affinities along the region's boundary, its perimeter, and texture difference with nearby pixels.  (Note that the authors trained their measure on PASCAL data, which is disjoint from ours.)  We stress that while this feature estimates how ``object-like'' a region is, it does not gauge importance.  It is useful for identifying full object segments, as opposed to fragments.

To model \textbf{object-like motion}, we develop a key-segments video segmentation descriptor~\cite{keysegments}.  It looks at the difference in motion patterns of a region relative to its closest surrounding regions.  Specifically, we compare optical flow histograms for the region and the pixels around it within a loosely fit bounding box.  Note that this feature is not simply looking for large motions or appearance changes from background.  Rather, we are describing how the motion of the region differs from its closest surrounding regions; this allows us to forgo assumptions about camera motion, and also to be sensitive to different magnitudes of motion.  Similar to the appearance feature above, it is useful for selecting object-like regions that ``stand-out'' from their surroundings.

To model the \textbf{likelihood of a person's face}, we compute the maximum overlap score $\frac{|q \cap r|}{|q \cup r|}$ between the region $r$ and any detected frontal face $q$ in the frame, using~\cite{viola}.

\textbf{\emph{Region features}:  }Finally, we compute the region's \textbf{size}, \textbf{centroid}, \textbf{bounding box centroid}, \textbf{bounding box width}, and \textbf{bounding box height}.  They reflect category-independent importance cues and are blind to the region's appearance or motion.  We expect that important people and objects will occur at non-random scales and locations in the frame, due to social and environmental factors that constrain their relative positioning to the camera wearer (e.g., sitting across a table from someone when having lunch, or handling cooking utensils at arm's length).  Our region features capture these statistics.

Altogether, these cues form a 14-dimensional feature space to describe each candidate region (4 egocentric, 3 object, and 7 region feature dimensions).

\subsubsection{Regressor to predict region importance}
\label{subsubsec:model}

Using the features defined above, we next train a model that can predict a region's importance.  The model should be able to learn and predict a region's \emph{degree} of importance instead of whether it is simply ``important'' or ``not important'', so that we can meaningfully adjust the compactness of the final summary (as we demonstrate in Section~\ref{sec:results}).  Thus, we opt to train a regressor rather than a classifier.

While the features defined above can be individually meaningful, we also expect significant interactions between the features.  For example, a region that is near the camera wearer's hand might be important only if it is also object-like in appearance.  Therefore, we train a linear regression model with pair-wise interaction terms to predict a region $r$'s \emph{importance score}:
\begin{equation}
I(r) = \beta_0 + \sum_{i=1}^{N} \beta_i x_i(r) + \sum_{i=1}^N \sum_{j=i+1}^N \beta_{i,j} x_i(r) x_j(r),
\label{eqn:model}
\end{equation}
where the $\beta$'s are the learned parameters, $x_i(r)$ is the $i$th feature value, and $N=14$ is the total number of features.

For training, we define a region $r$'s target importance score by its maximum overlap $\frac{|GT \cap r|}{|GT \cup r|}$ with any ground-truth region $GT$ in a training video obtained from Section~\ref{subsec:mturk}.  Thus, regions with perfect overlap with ground-truth will have a target importance score of $1$, those with no overlap with ground-truth will have an importance score of $0$, and all others will have an importance score in $(0,1)$.  We standardize the features to zero-mean and unit-variance, and solve for the $\beta$'s using least-squares.  For testing, our model takes as input a region $r$'s features (the $x_i$'s) and predicts its importance score $I(r)$.  Note that we train and test using video from different users to avoid overfitting our model to any specific camera wearer.

\subsection{Segmenting the video into temporal events}
\label{subsec:events}

Given a new video, we first partition the video temporally into events, and then isolate the important people and objects in each event.  Events allow the final summary to include multiple instances of an object/person that is central in multiple contexts in the video.  For example, suppose that the camera wearer plays with her dog at home in the morning and later takes the dog out to the park at night.  We can treat the two instances of the dog as different objects (since they appear in different events) and include both in the final summary.  Moreover, events indicate which selected frames are more related to one another, giving a hierarchical structure to the final summary.

While shot boundary detection has been frequently used to perform event segmentation for videos, it is impractical for our wearable camera data setting. Traditional shot detection generally assumes visual continuity and thus tends to oversegment egocentric events due to frequent head movements.  Instead, we detect egocentric events by clustering scenes in such a way that frames with similar global appearance can be grouped together even when there are a few unrelated frames (``gaps") between them.

Let $\mathcal{V}$ denote the set of all video frames.  We compute a pairwise distance matrix $D_\mathcal{V}$ between all frames $f_m,f_n \in \mathcal{V}$, using the distance:
\begin{equation}
D(f_m, f_n) = 1 - w^t_{m,n} \exp\Bigl(-\frac{1}{\Omega}\chi^2(f_m, f_n)\Bigr),
\label{eqn:frameColorAffinity}
\end{equation}
where $w^t_{m,n} = \frac{1}{t}\max(0, t-|m-n|)$, $t$ is the size of the temporal window surrounding frame $f_m$, $\chi^2(f_m, f_n)$ is the $\chi^2$-distance between color histograms of $f_m$ and $f_n$, and $\Omega$ denotes the mean of the $\chi^2$-distances among all frames.  Thus, frames similar in color receive a low distance, subject to a weight that discourages frames too distant in time from being grouped.

\begin{figure}[t]
\centering
\hspace*{-0.1in}
\includegraphics[width=8.7cm]{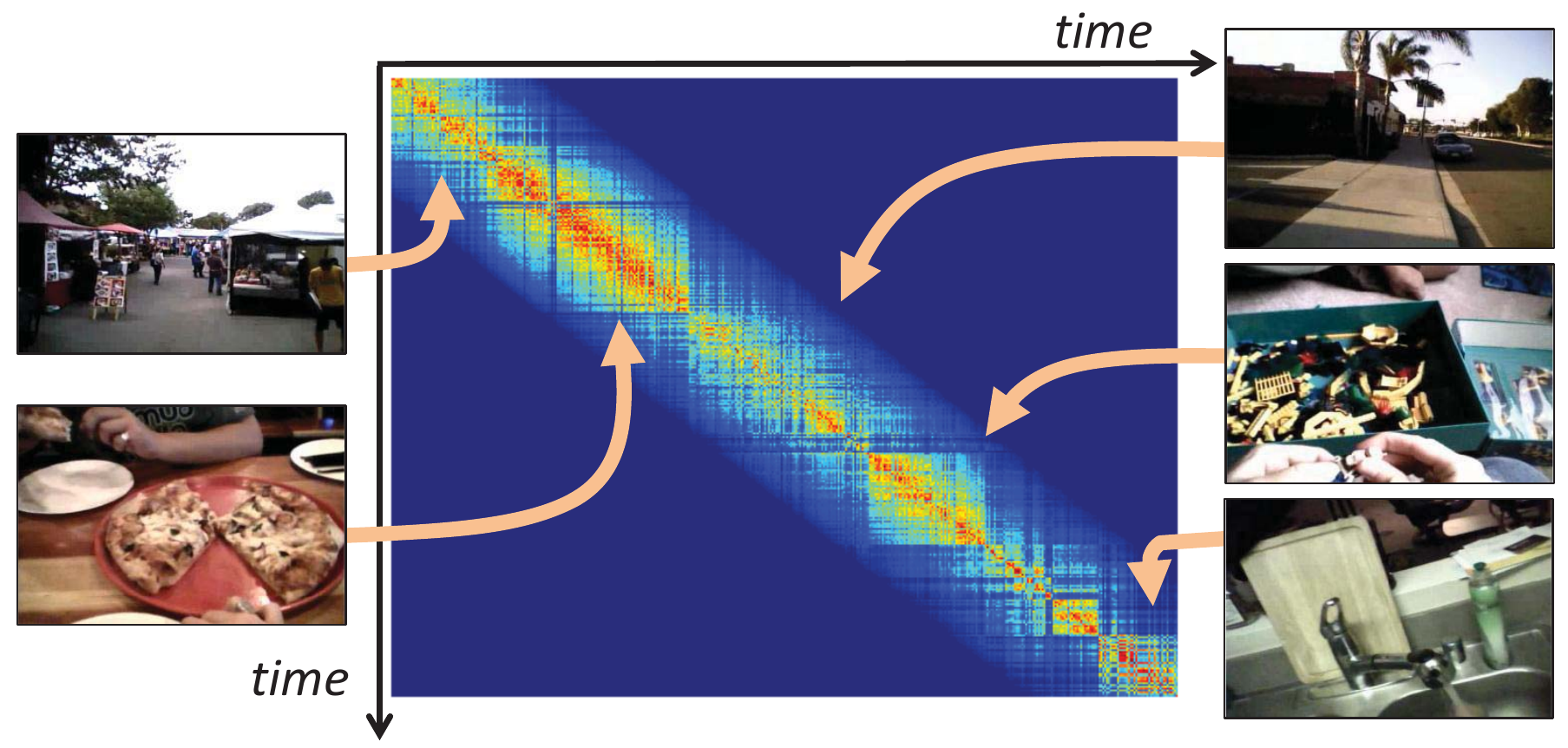}
\caption{Distance matrix that measures global color dissimilarity between all frames.  (Blue/red reflects high/low distance.)  The images show representative frames of each discovered event.  The block structure along the diagonal reveals groups of frames that are close in appearance and time.}
\label{fig:dmatrix}
\end{figure}

We next perform complete-link agglomerative clustering with $D_\mathcal{V}$, grouping frames until the smallest maximum inter-frame distance is larger than two standard deviations beyond $\Omega$.  The first and last frames in a cluster determine the start and end frames of an event, respectively. Fig.~\ref{fig:dmatrix} shows the distance matrix computed for one subject's day, and the representative frames for each discovered event.

\subsection{Discovering an event's key people/objects}
\label{subsec:regionClustering}

For each event, we aim to select the important people and objects that will go into the final summary, while avoiding redundancy.  Recall that objects are represented at the frame-level (Section~\ref{subsec:learning}).  Thus, our goal is to group together instances of the same person or object that appear over time in each event.

Given an event, we first score each bottom-up segment in each frame using our regressor.  Since we do not know a priori how many important things an event contains, we generate a candidate pool of clusters from the set $\mathcal{C}$ of bottom-up regions, and then remove any redundant clusters, as follows.

\begin{figure}[t]
\centering
\includegraphics[width=8cm]{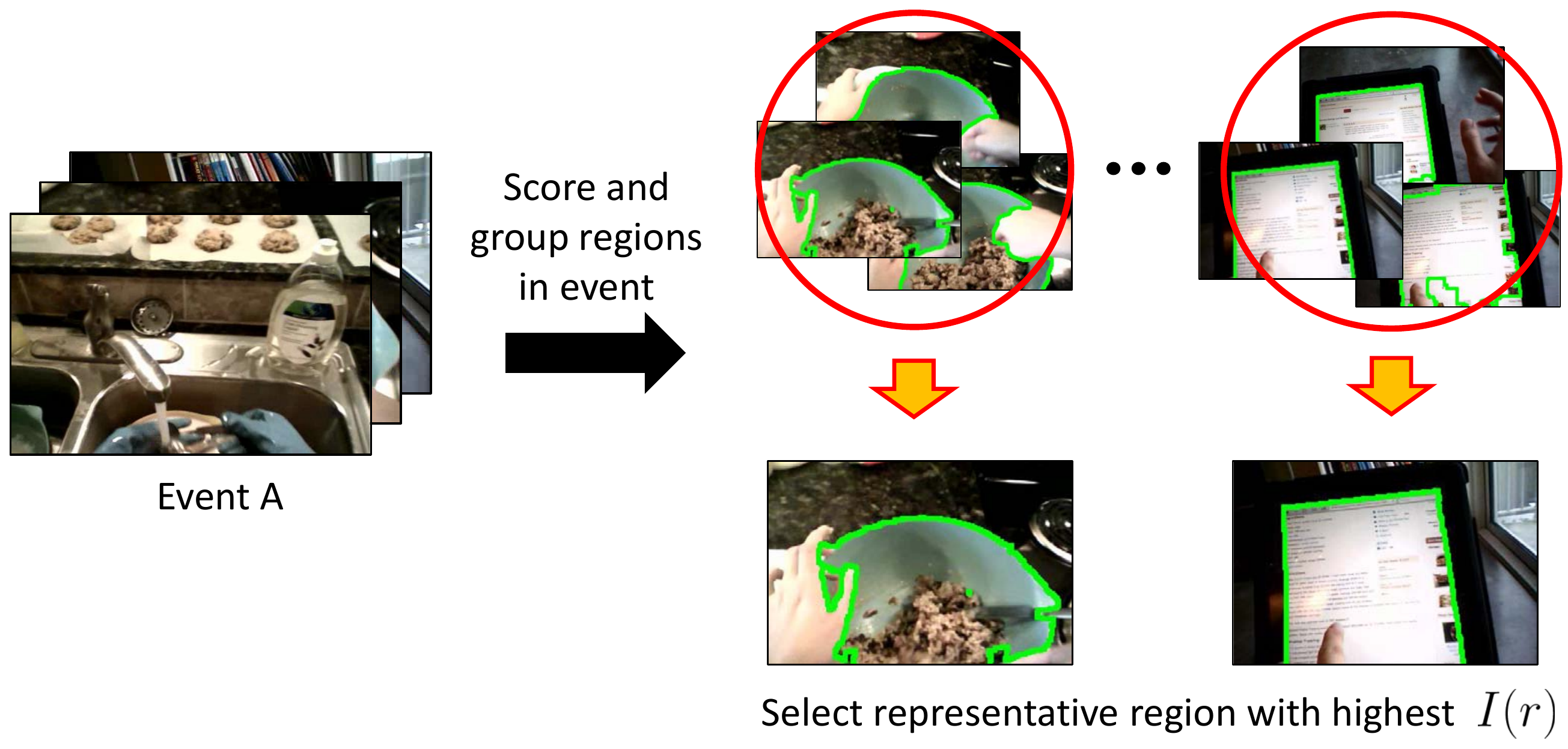}
\caption{Discovering an event's key people and objects.  For each event, we group together regions that are likely to belong to the same object, and then for each group, we select the region with the highest importance score as its representative.}
\label{fig:regionGrouping}
\end{figure}

To extract the candidate groups, we first compute an affinity matrix $K_\mathcal{C}$ over all pairs of regions $r_m,r_n \in \mathcal{C}$, where affinity is determined by color similarity: $K_\mathcal{C}(r_m, r_n) = \exp(-\frac{1}{\Gamma}\chi^2(r_m, r_n))$, where $\Gamma$ denotes the mean $\chi^2$-distance among all pairs in $\mathcal{C}$.  We next partition $K_\mathcal{C}$ into multiple (possibly overlapping) inlier/outlier clusters using a factorization approach~\cite{perona-eccv1998}.  The method finds tight sub-graphs within the input affinity graph while resisting the influence of outliers.  Each resulting sub-graph consists of a candidate important object's instances.  To reduce redundancy, we sort the sub-graph clusters by the average $I(r)$ of their member regions, and remove those with high affinity to a higher-ranked cluster.  Finally, for each remaining cluster, we select the region with the highest importance score as its representative (see Fig.~\ref{fig:regionGrouping}).

\subsection{Generating a storyboard summary}
\label{subsec:storyboard}

Finally, we create a storyboard visual summary of the video.  We display the event boundaries and frames of the selected important people and objects (see Fig.~\ref{fig:qualitative}).  Each event can display a varying number of frames, depending on how many unique important things our method discovers.

We propose two ways to adjust the \emph{compactness} of the summary: (1) according to a target importance criterion, and (2) according to a target summary length.  We describe each process in detail next.

\subsubsection{Summarization given an importance criterion}
\label{subsec:automatic}

We first describe how to summarize the video given an importance criterion.  This allows the system to automatically produce the most compact summary possible that encapsulates only the people and objects that meet the importance threshold.

When discovering an event's key people and objects (Section~\ref{subsec:regionClustering}), we take only those regions that have importance scores higher than the specified criterion to form set $\mathcal{C}$.  We then proceed to group instances of the same person or object together in $\mathcal{C}$, and select the frame with the highest scoring region in each group to go into the summary.

\subsubsection{Summarization given a length budget} 
\label{subsec:dp}

Alternatively, we can summarize the video given a length budget $k$.  This allows the system to answer requests such as, ``Generate a 5-minute summary.''  We formulate the objective as a \emph{k}-frame selection problem and define the following energy function:
\begin{align}
E(S) = - & \sum_{i=1}^{|S|} I(f_{s_i}) + \sum_{i=1}^{|S|-1} \exp\Bigl(-\frac{1}{\Omega} \chi^2(f_{s_i},f_{s_{i+1}})\Bigr)  \notag \\
- & \sum_{i=1}^{|S|-1} |s_i - s_{i+1}|^\frac{1}{2},
\label{eqn:energy}
\end{align}
where $S = \{s_1, \dots, s_k\}$ is the set of indices of the $k$ selected frames, and $\Omega$ is the mean of the $\chi^2$-distances among all frames.

There are three terms in our energy function.  The first term enforces selection of important frames, since we want the summary to contain the discovered important people and objects.  We score each frame using the region that has the highest importance score: $I(f_{s_i}) = \max_m I(r_{m,i})$, where $r_{m,i}$ is the $m$th region in frame $i$.  Our second term enforces visual uniqueness, i.e., that adjacent selected frames contain different objects.  We want the summary to avoid including redundant frames.  Thus, we compute an affinity based on the $\chi^2$-distance between color histograms of adjacent frames $f_{s_i}$ and $f_{s_{i+1}}$.  Finally, our last term enforces selection of frames that are spread out in time such that the summary best captures the entire ``story'' of the original video.  For this, we compute the difference in frame index of the selected frames.  Note that $\sum_{i=1}^{|S|-1} |s_i - s_{i+1}|^\frac{1}{2}$ achieves a maximum when the temporal distances between all adjacent frames $|s_i - s_{i+1}|$ are equal.

We compute the optimal set $S^*$ of $k$ frames by finding the set that minimizes Eqn.~\ref{eqn:energy}:
\begin{equation}
S^* = \underset{S \subset V}{\operatorname{argmin}} ~E(S),
\label{eqn:maxenergy}
\end{equation}
where $V$ is the set of frames of the selected important people and objects from Section~\ref{subsec:regionClustering}.

A naive approach for optimizing Eqn.~\ref{eqn:maxenergy} would take time $O({F\choose k})$ for $F=|V|$ total frames.  Instead, we efficiently find the optimal set $S^*$ using dynamic programming, by exploiting the optimal substructure that exists in the $k$-frame selection problem.  

Specifically, the minimum energy $M(f_n,t)$ of a $t$-length summary that selects frame $f_n$ at time step $t$ can be recursively computed as follows:
\small
\begin{align}
M(f_n,t)_{t \le n \le F-k+t}  = ~~~~~~~~~~~~~~~~~~~~~~~~~~~~~~~~~~~~~~~~~~~~~~~~~~~ & \notag \\
\notag \\
    \begin{cases}
        -I(f_n), & \text{if $t = 1$}.\\
        -I(f_n) + \displaystyle\min_{p \le m \le q}( e(f_m,f_n) + M(f_m, t-1) ), & \text{if $1 < t \le k$},\\
    \end{cases}
\label{eqn:recursive}
\end{align}
\normalsize
where $p = t-1$, $q = F-k+t+1$, and $e(f_m,f_n) = \exp(-\frac{1}{\Omega} \chi^2(f_m,f_n)) - |m - n|^\frac{1}{2}$.  We enforce the selected set of frames to be a temporally ordered subsequence of the original video: $s_i < s_{i+1}, \forall i$.  Thus, any ``path'' that does not obey this rule is assigned infinite cost.  

Using Eqn.~\ref{eqn:recursive}, we can compute the minimum energy for a $k$-length summary as $E(S^*) = \min_n M(f_n,k)$, which can be solved in $O(F^2k)$ time.  We retrieve the optimal set of $k$ frames $S^*$ by backtracking from $f_n$ at time $k$.

\subsubsection{Discussion}
\label{subsec:discussion}

The two strategies presented above offer certain trade-offs.  The importance criterion automatically produces the most compact summary possible that includes all unique instances of the important people and objects; however, it does not give the user direct control on the length of the output summary.  In contrast, while the proposed budgeted formulation can return a storyboard of a specified length, it does not permit setting an absolute threshold on how important objects must be for inclusion.

In addition to being a compact video diary of one's day, our storyboard summary can be considered as a \emph{visual index} to help a user peruse specific parts of the video.  This would be useful when one wants to relive a specific moment or search for less important people or objects that occurred with those found by our method.

Alg.~\ref{alg:egocentric} recaps all the steps of our approach.

\begin{algorithm}[t]
    \caption{Our summarization approach}
    \SetLine
    \KwIn{Egocentric video, and importance selection criterion or length budget $k$.}
    \KwOut{Storyboard summary.}

    1.  Train regression model.  (Sec.~\ref{subsec:learning})\\
    2.  Segment video into temporal events.  (Sec.~\ref{subsec:events})\\
    \textit{For each event,}\\
    3.  Compute $I(r)$ for all regions.  (Sec.~\ref{subsec:learning})\\
    4.  Group regions that belong to same person/object.  (Sec.~\ref{subsec:regionClustering})\\
    5.  Retain unique clusters, select most important region in each group. (Sec.~\ref{subsec:regionClustering})\\
    6.  Generate storyboard summary that shows selected important people/objects.  (Sec.~\ref{subsec:storyboard})
    \label{alg:egocentric}
\end{algorithm}

\section{Results}
\label{sec:results}

\yj{In this section we evaluate our approach on our new UT Egocentric (UT Ego) dataset and on the Activities of Daily Living (ADL) dataset~\cite{deva-egocentric-cvpr2012}, which consists of 17 and 10 hours of egocentric video, respectively.  We offer direct comparisons to existing methods for both saliency and video summarization, and we perform a user study with over 25 subjects to quantify the perceived quality of our results.  We use UT Ego for the experiments in Sections~\ref{subsec:imp_accuracy}-6, and ADL for the experiments in Section~\ref{subsec:adl}.}

\begin{figure}[t]
\centering
\includegraphics[width=6cm]{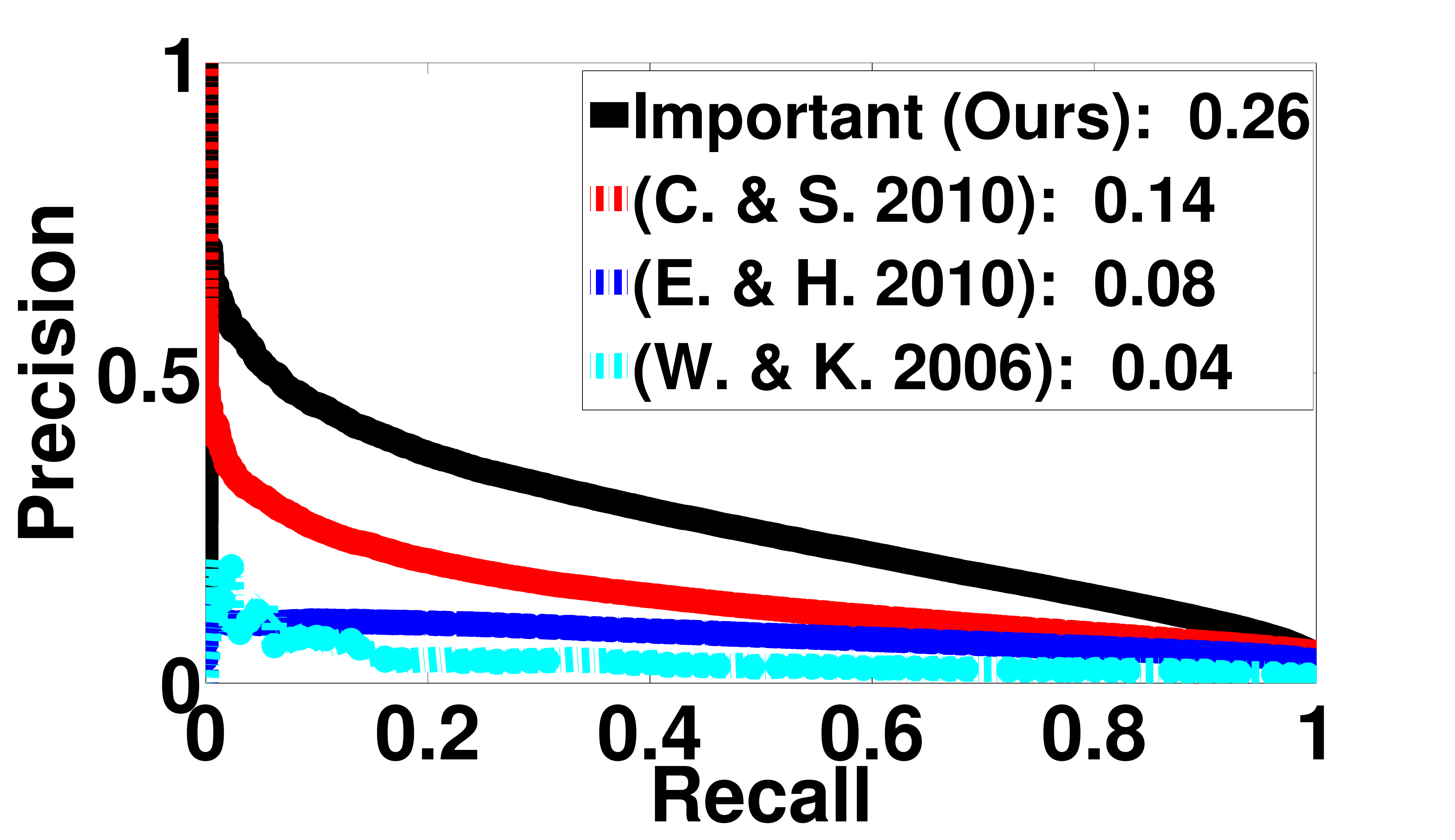}
\caption{Precision-Recall for important object prediction.  Numbers in the legends denote average precision.  By leveraging egocentric-specific cues, our approach more accurately discovers the important regions.}
\label{fig:importancePR} 
\end{figure}

\begin{figure*}[t]
\centering
\includegraphics[width=16cm]{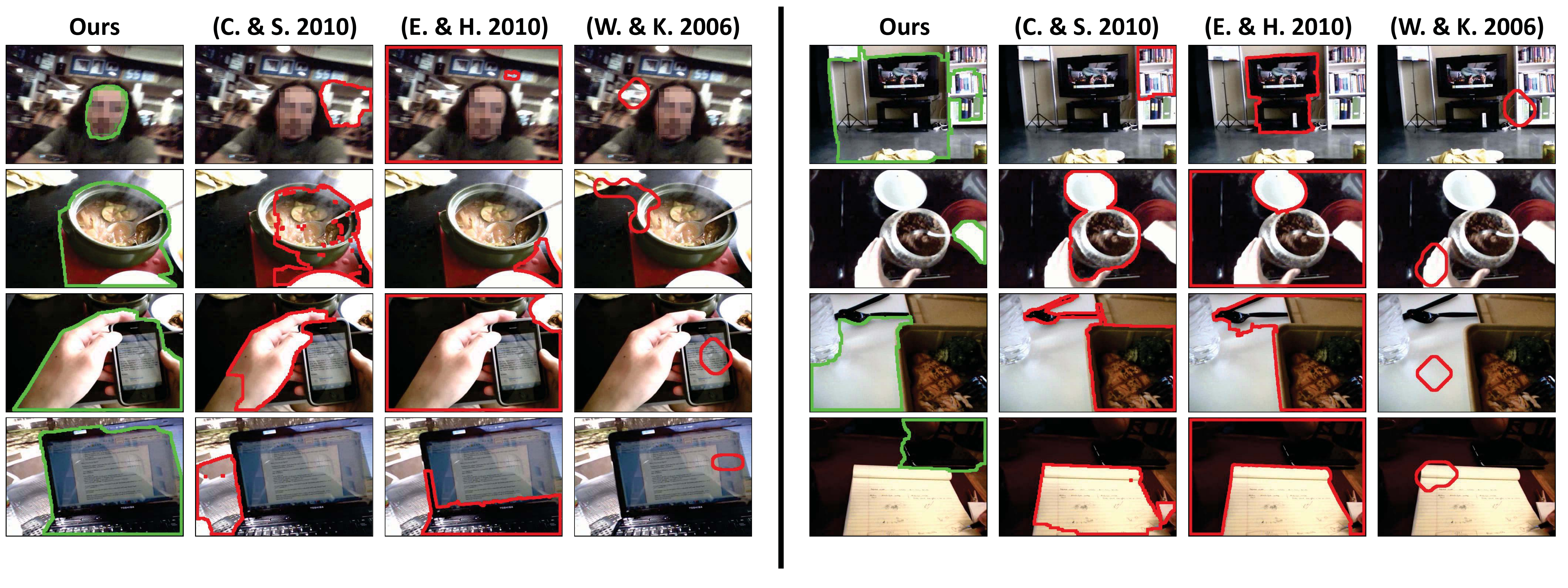}
\caption{Example selected regions/frames.  The first four columns show examples of correct predictions made by our approach, and the last four columns show failure cases in which the high-level saliency methods~\cite{carreira-mincut,endres-eccv2010} make better predictions.}
\label{fig:importanceExamples}
\end{figure*}

\subsection{Dataset and implementation details}

For our UT Ego dataset, we collected 10 videos from four subjects, each 3-5 hours long.$^1$  Each person contributed one video, except one who contributed seven.  The videos are challenging due to frequent camera viewpoint/illumination changes and motion blur.  For evaluation, we use four data splits: for each split we train with data from three users and test on one video from the remaining user.  Hence, the camera wearers in any given training set are disjoint from those in the test set, ensuring we do not learn user- or object-specific cues.

\yj{ADL contains 20 videos from chest-mounted cameras, each on average about 30 minutes long.  The camera wearers perform daily activities in the house, like brushing hair, cooking, washing dishes, or watching TV.  To generate candidate object regions on ADL, we use BING~\cite{bing}, which generates bounding box proposals and is orders of magnitude faster than the min-cut approach of~\cite{carreira-mincut}.}

We use Lab space color histograms, with 23 bins per channel, and optical flow histograms with 61 bins per direction using~\cite{brox-pami2011}.  We set $t=27000$ and $t=2250$ (i.e., a 60 and 5 minute temporal window), for UT Ego and ADL, respectively.  We set $\theta_r=10000$ and $\theta_p = 0.7$ after visually examining a few examples.  We fix all parameters for all results.  For efficiency, we process every 15th frame (i.e., 1 fps).  For Eqn.~\ref{eqn:energy}, we standardize each term to zero-mean and unit-variance using training data.

\subsection{Important region prediction accuracy}
\label{subsec:imp_accuracy}

We first evaluate our method's ability to predict important regions, compared to three state-of-the-art methods: (1) the object-like score of~\cite{carreira-mincut}, (2) the object-like score of~\cite{endres-eccv2010}, and (3) a bottom-up saliency detector~\cite{walther-nn2006}.  The first two are high-level learned functions that predict a region's likelihood of overlapping a true object, whereas the third is a low-level detector to find regions that ``stand-out''.  They are all general-purpose metrics (not tailored to egocentric data), so they allow us to gauge the impact of our proposed egocentric cues for finding important objects in video.

We use the annotations obtained on MTurk as ground truth (GT) (see Sec.~\ref{subsec:mturk}).  Some frames contain more than one important region, and some contain none, depending on what the annotators deemed important.  On average, each video contains 680 annotated frames and 280,000 test regions.  A region $r$ is considered to be a true positive (i.e., important object), if its overlap score with any GT region is greater than 0.5, following PASCAL convention.

Fig.~\ref{fig:importancePR} shows precision-recall curves on all test regions across all train/test splits.  Our approach predicts important regions significantly better than all three existing methods.  The two high-level methods~\cite{carreira-mincut,endres-eccv2010} can successfully find prominent object-like regions, and so they noticeably outperform the low-level saliency detector.  However, by focusing on detecting \emph{any} object, unlike our approach they are unable to distinguish those that may be important to a camera wearer.

Fig.~\ref{fig:importanceExamples} shows example important regions detected by each method.  The first four columns show examples of correct predictions made by our method.  We see that low-level saliency detection~\cite{walther-nn2006} is insufficient; its local estimates fail to find object-like regions.  For example, it finds a bright blob surrounded by a dark region to be the most salient (first row, fourth column).

\begin{figure}[t]
\centering
\includegraphics[width=8.5cm]{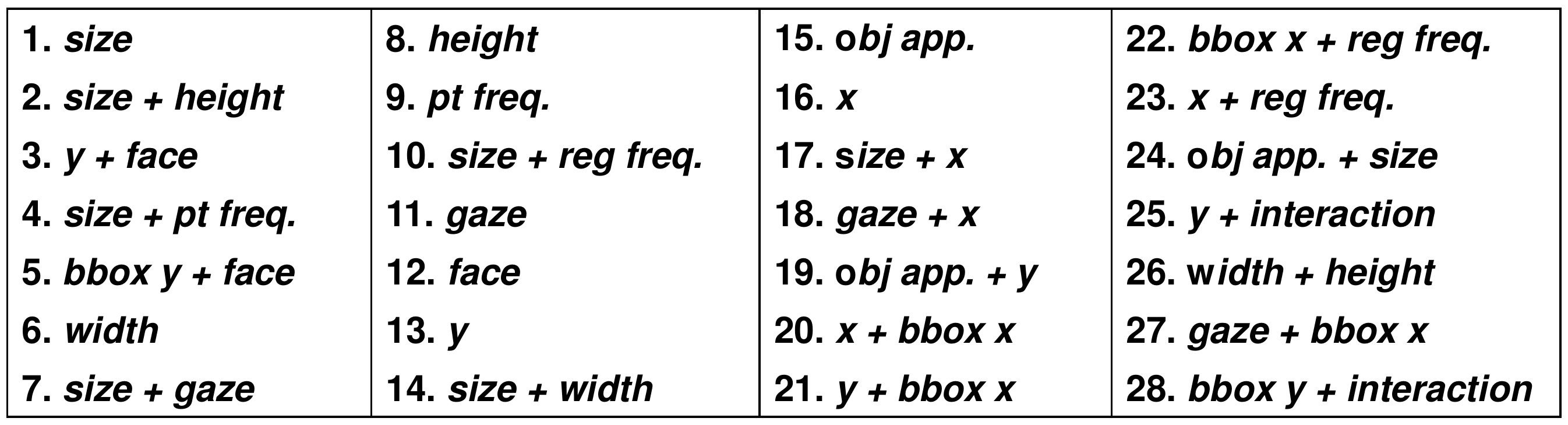}
\caption{Top 28 features with highest learned weights.}
\label{fig:importantCues}
\end{figure}

The last four columns show examples of incorrect predictions made by our method.  The high-level saliency detection methods~\cite{carreira-mincut,endres-eccv2010} produce better predictions for these examples.  In the first example, our method produces an under-segmentation of the important object and includes regions surrounding the television due to the combined region having higher object-like appearance score than the television alone.  In the second example, our method incorrectly detects the user's hand to be important, while in the third and fourth examples, it determines background regions to be important due to their high frequency.  

\yj{We next perform ablation studies to investigate the contribution of the pairwise interaction terms of our importance predictor.  Specifically, we compare to a linear regression model and an L1-regularized linear regression model using only the original 14-dimensional features.  The average precision of the linear regression model is 0.20, and the average precision of the L1-regularized model ranges from 0.14-0.20 depending on the level of sparsity, as enforced by the weight on the regularization term.  This result shows that the original features alone are not sufficiently expressive, and that the pairwise terms are necessary to more fully capture the relationship between the features and desired importance values.}

\subsection{Which cues matter most for importance?}%

Fig.~\ref{fig:importantCues} shows the top 28 out of 105 $\bigl(= 14 + {14 \choose 2}\bigr)$ features that receive the highest learned weights (i.e., $\beta$ magnitudes).  Region size is the highest weighted cue, which is reasonable since an important person/object is likely to appear roughly at a fixed distance from the camera wearer.  Among the egocentric features, gaze and frequency have the highest weights.  Frontal face overlap is also highly weighted; intuitively, an important person would likely be facing and conversing with the camera wearer.

Some highly weighted pair-wise interaction terms are also quite interesting.  The feature measuring a region's face overlap \emph{and} y-position has more impact on importance than face overlap alone.  This suggests that an important person usually appears at a fixed height relative to the camera wearer.  Similarly, the feature for object-like appearance \emph{and} y-position has high weight, suggesting that a camera wearer often adjusts his ego-frame of reference to view an important object at a particular height.

Surprisingly, the pairing of the interaction (distance to hand) and frequency cues receives the lowest weight.  A plausible explanation is that the \emph{frequency} of a handled object highly depends on the camera wearer's activity.  For example, when eating, the camera wearer's hand will be visible and the food will appear frequently.  On the other hand, when grocery shopping, the important item s/he grabs from the shelf will (likely) be seen for only a short time.  These conflicting signals would lead to this pair-wise term having low weight.  Another paired term with low weight is an ``object-like'' region that is frequent; this is likely due to unimportant background objects (e.g., the lamp behind the camera wearer's companion).  This suggests that higher-order terms could yield even more informative features.

\subsection{Importance-based summarization accuracy}
\label{subsec:summarizationAccuracy}

Next we evaluate our method's summarization results using the importance-based criterion, and in the following section we evaluate its budget-based results.

\subsubsection{Quantitative evaluation}
\label{subsubsec:quantEval1}

The central premise of our work is that day-to-day activity viewed from the first person perspective largely revolves around the important people and objects with which the camera wearer interacts.  Accordingly, a good visual summary must capture those important entities.  Thus, we analyze the recall rate for our method and two competing summarization strategies.  The first is uniform keyframe sampling, and the second is event-based adaptive keyframe sampling.  The latter computes events using the same procedure as our method (Sec.~\ref{subsec:events}), and then divides its keyframes evenly across events.  Both methods are modeled after standard keyframe and event detection methods~\cite{summary-survey-2008,wolf-1996,zhang-pr1997}.

\begin{figure*}[th!]
\centering
\begin{tabular}{cccc}
\includegraphics[width=4.3cm]{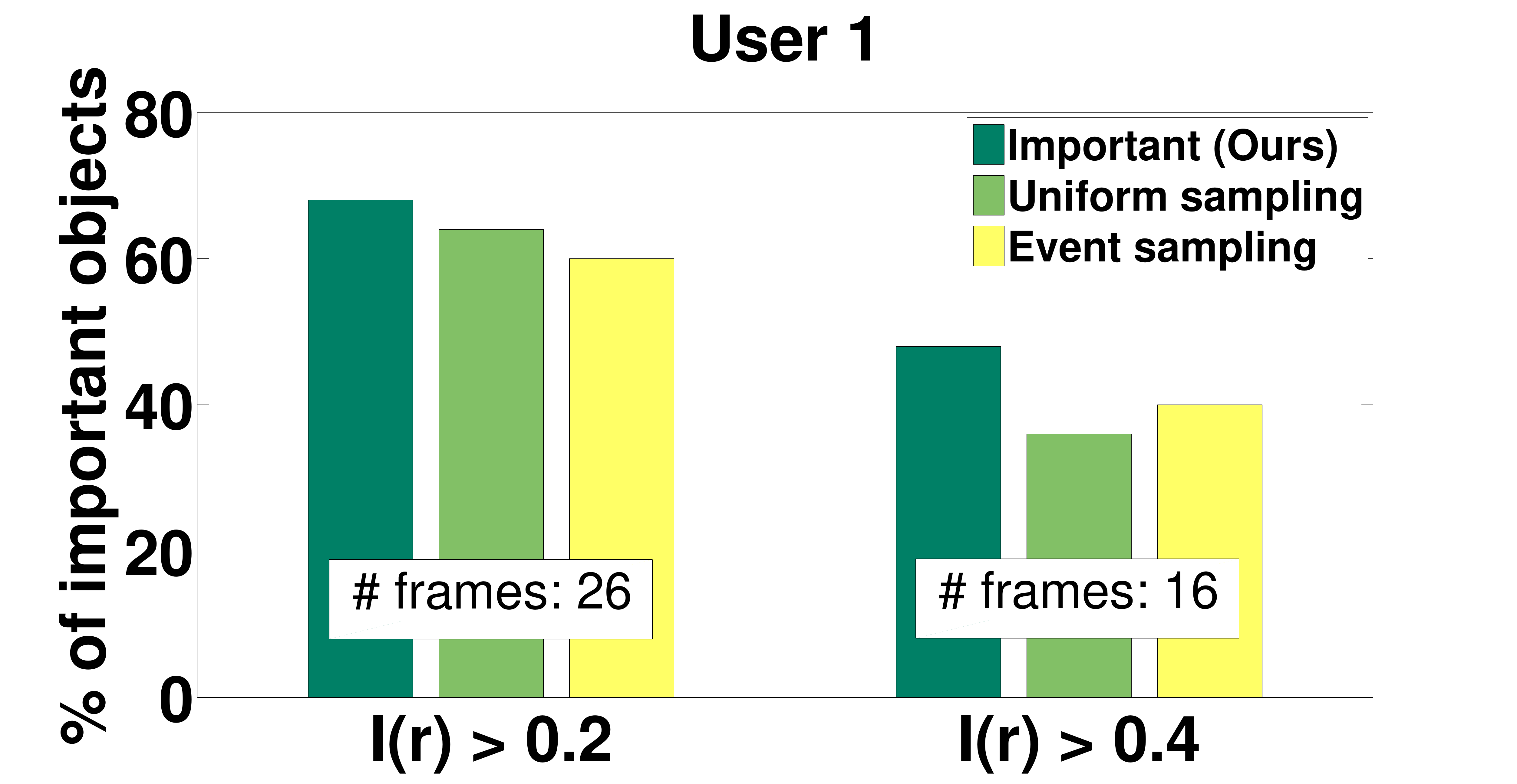}
\includegraphics[width=4.3cm]{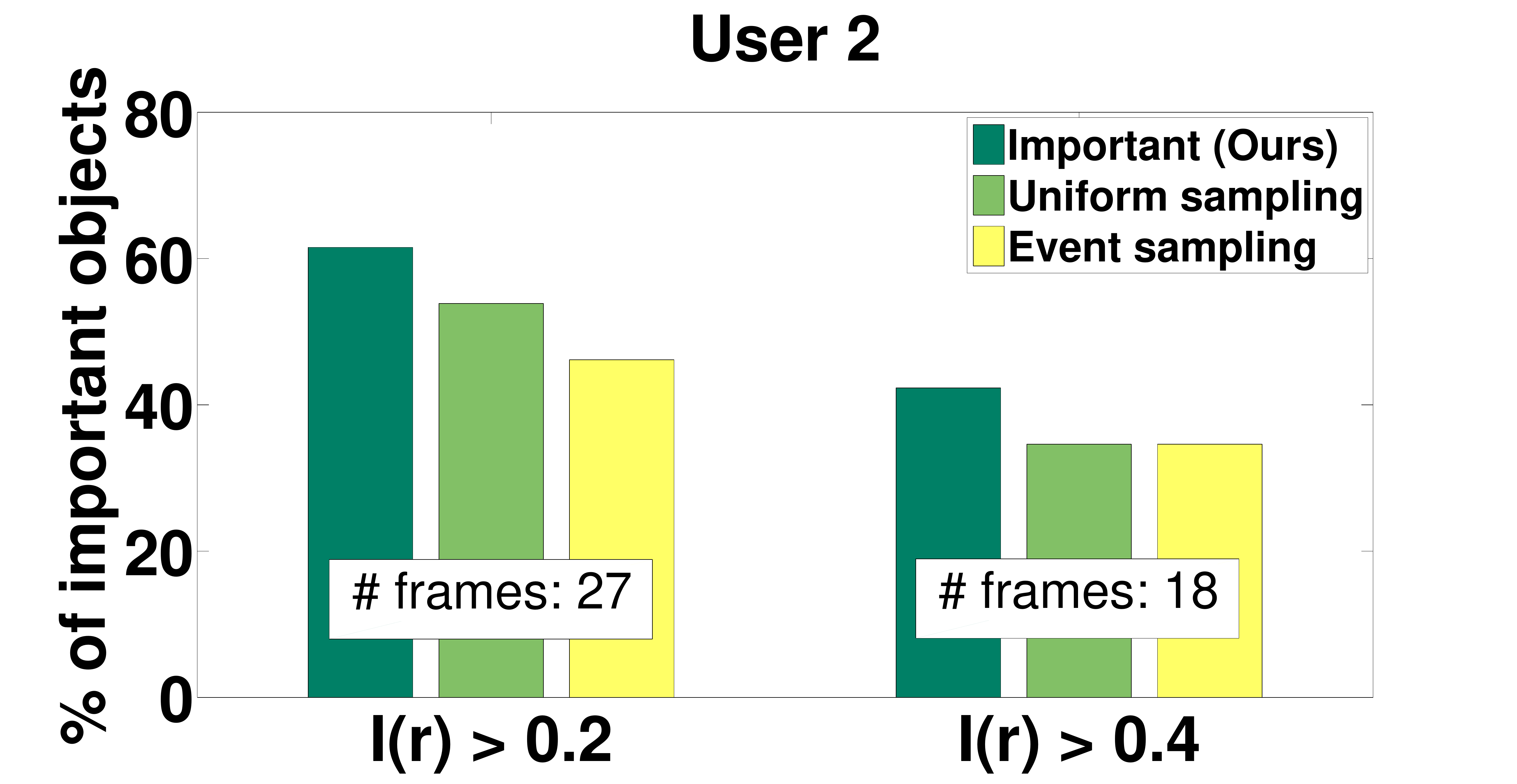}
\includegraphics[width=4.3cm]{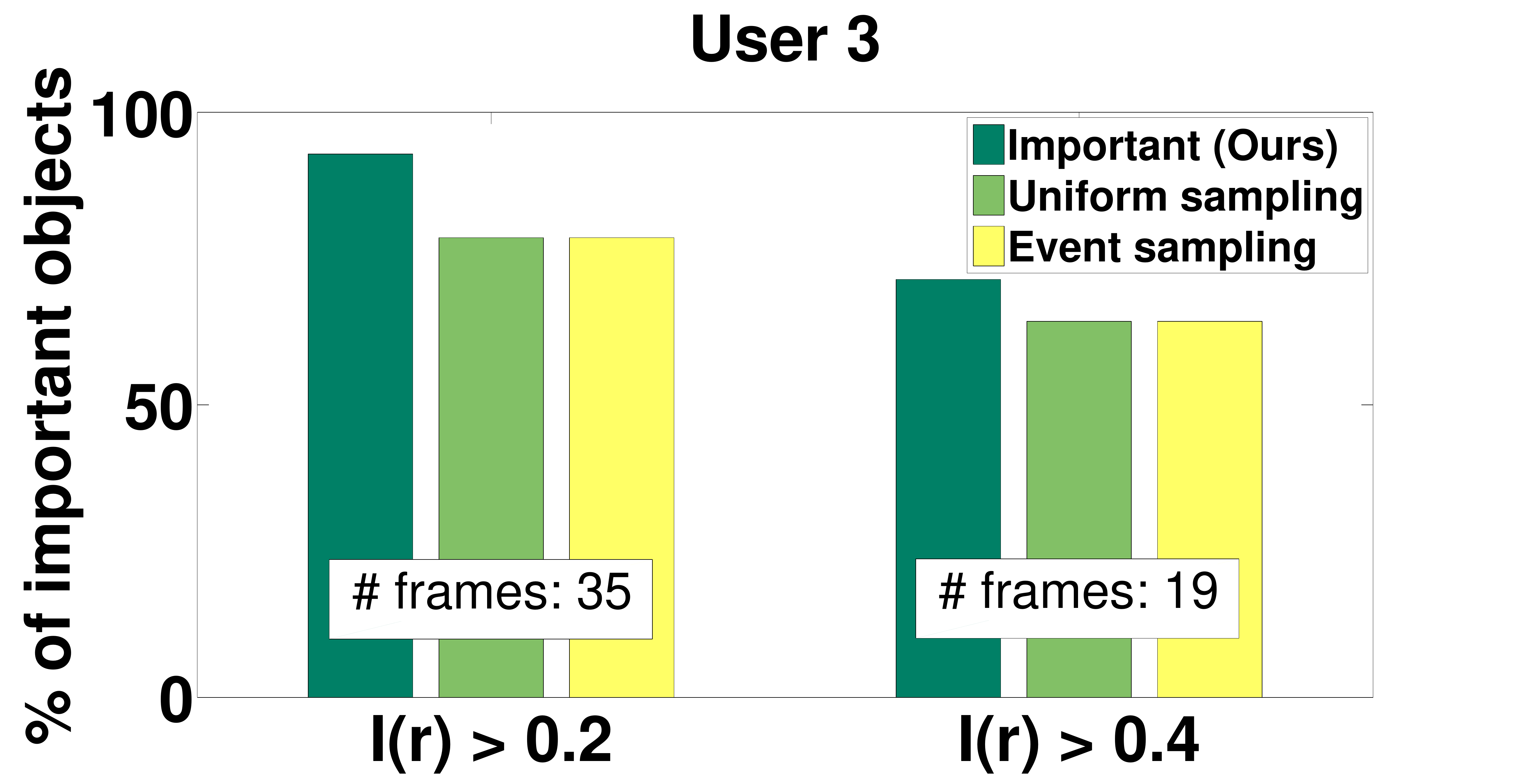}
\includegraphics[width=4.3cm]{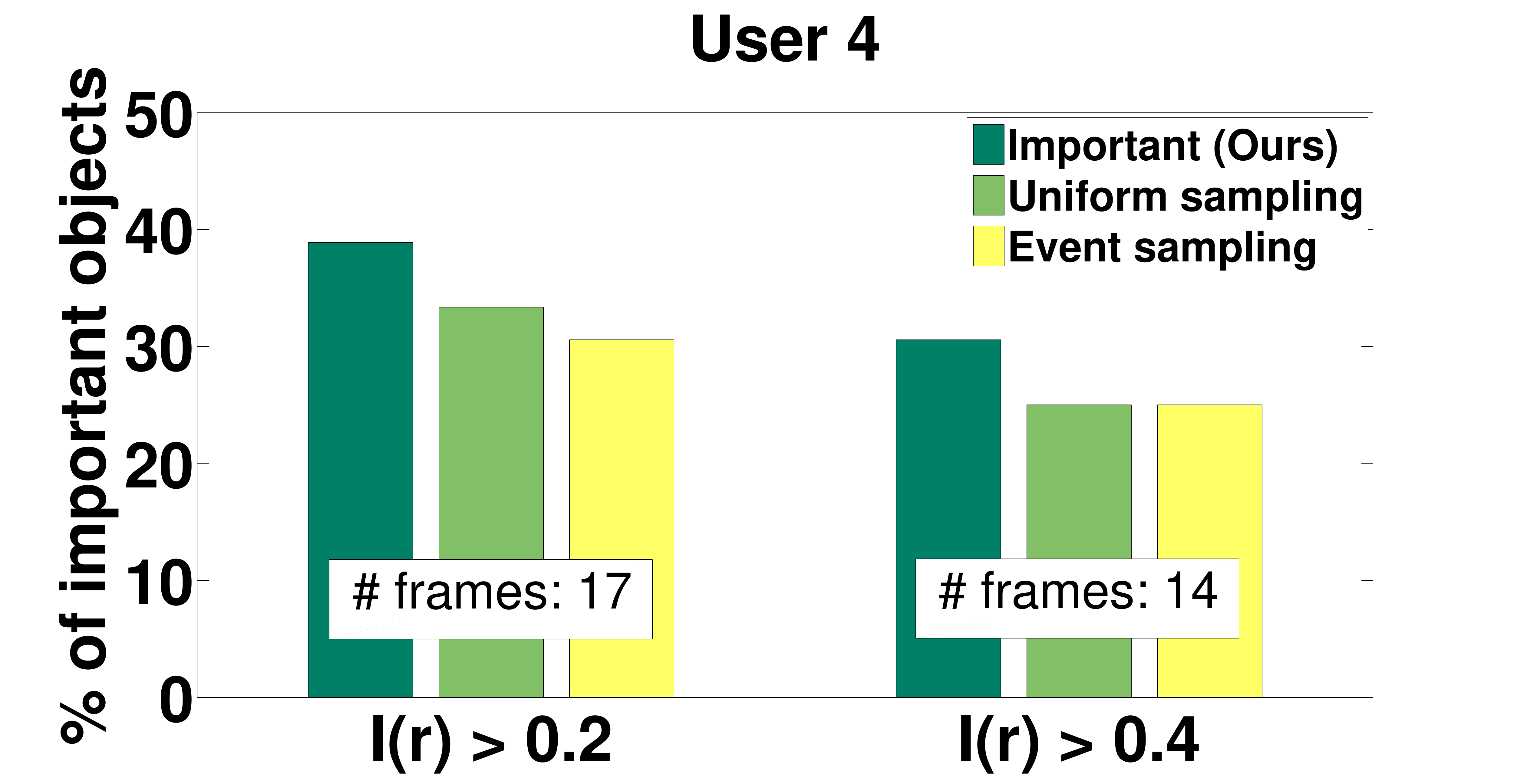}
\end{tabular}
\caption{Comparison to alternative summarization strategies, in terms of important object recall rate.  Using the same number of frames, our approach includes more important people and objects.}
\label{fig:summarizationResults}  %
\end{figure*}

Fig.~\ref{fig:summarizationResults} shows the results.  Each set of bars shows the recall rates for the three methods.  Our method varies its selection criterion on $I(r)$ over $\{0.2, 0.4\}$, for two summaries in total for each user.  These thresholds are used to cover a broad spectrum (i.e., low and high selection criteria) and are arbitrary; we see consistent relative results for any threshold.  To compare our recall rates to those of the baselines, we create summaries for the baselines with the same number of frames as ours.

\begin{figure}[t!]
\centering
\hspace*{-0.15in}
\begin{tabular}{cc}
\includegraphics[width=5.1cm]{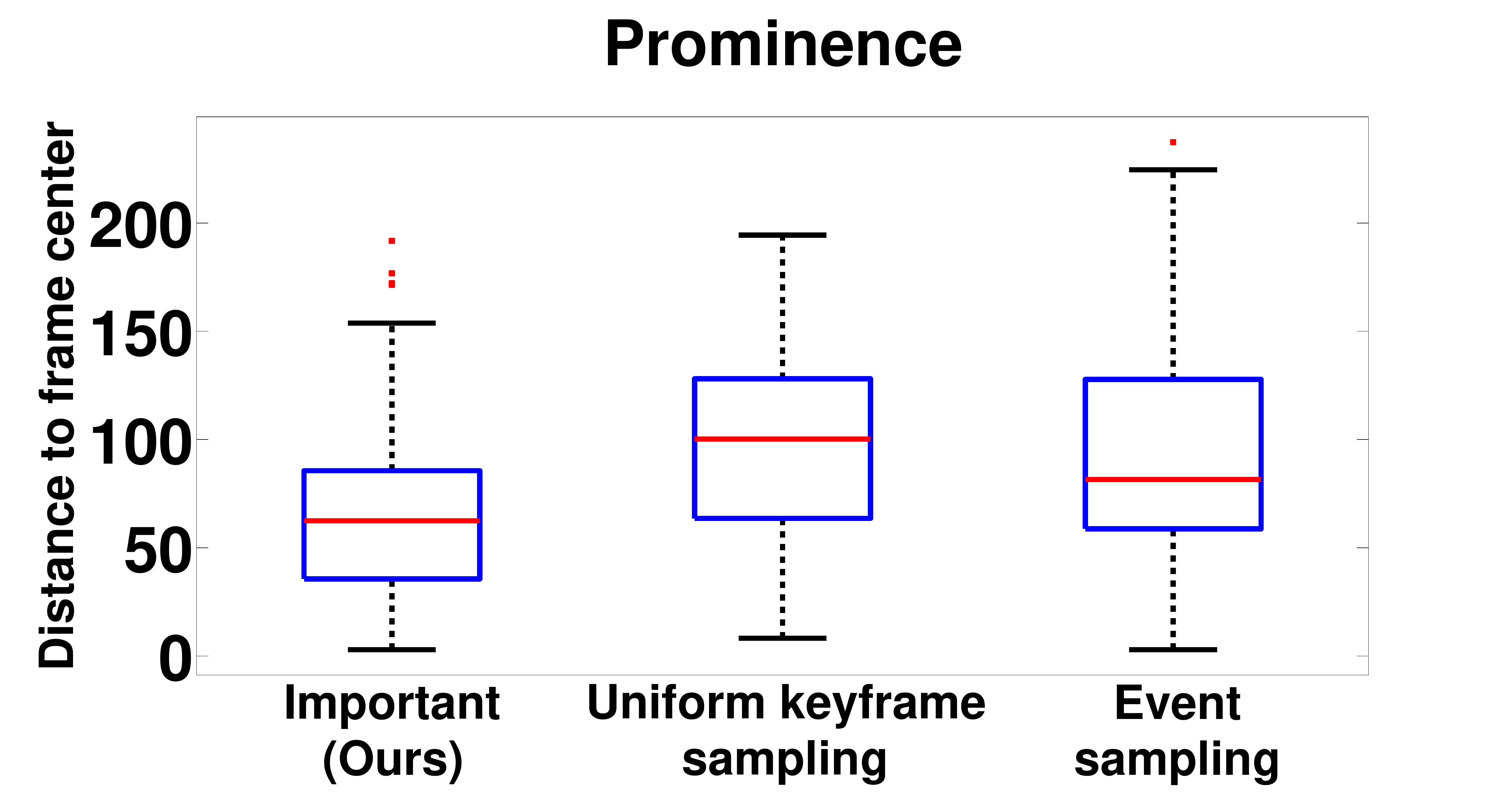}
\includegraphics[width=3.3cm]{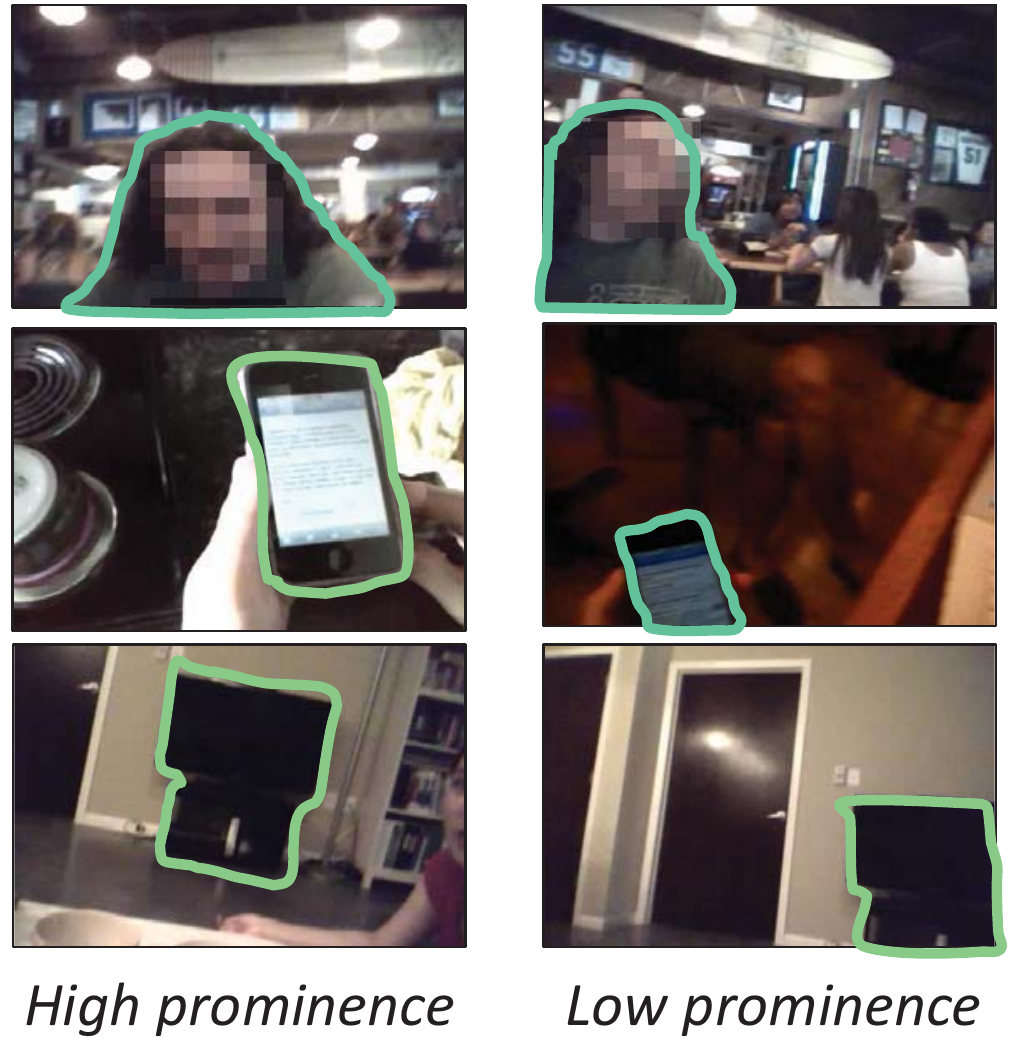}
\end{tabular}
\caption{Comparison to alternative summarization strategies, in terms of the prominence of the objects within selected keyframes. Our summaries more prominently display the important objects.}
\label{fig:prominence}
\end{figure}

If a frame contains multiple important objects, we score only the main one.  Likewise, if a summary contains multiple instances of the same GT object, it gets credit only once.  Note that this measure is favorable to the baselines, since it does not consider object \emph{prominence} in the frame.  For example, we give credit for the TV in the last frame in Fig.~\ref{fig:prominence}, bottom row, even though it is only partially captured.  Furthermore, by definition, the uniform and event-based baselines are likely to get many hits for the most frequent objects.  These make the baselines very strong and meaningful comparisons.

\begin{figure*}[t!]
\centering
\includegraphics[width=17cm]{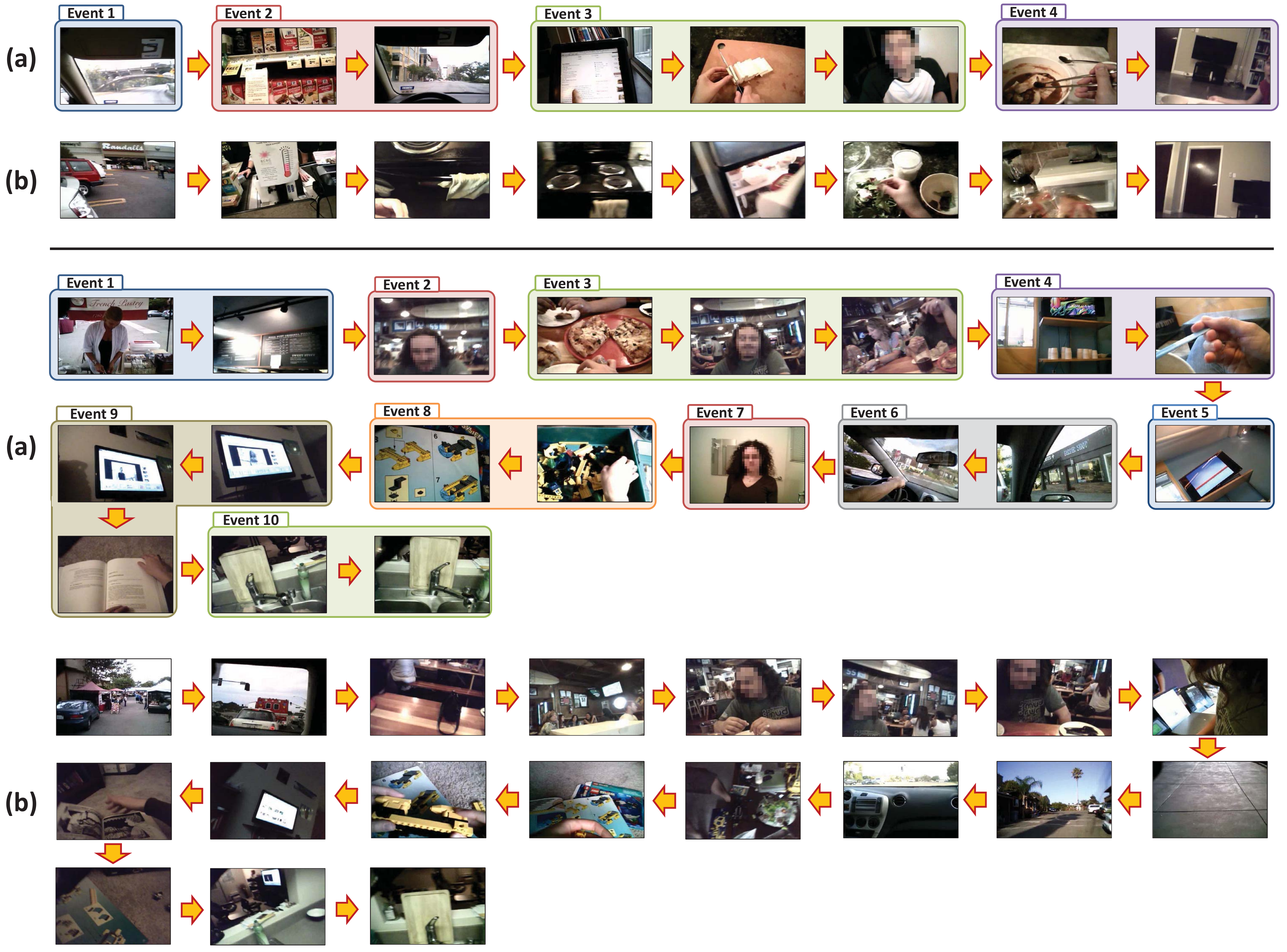}
\caption{(a) Our summary versus (b) uniform keyframe sampling. The colored blocks for ours indicate the discovered events.  Our summary focuses on the important people and objects.  While uniform keyframe sampling does hint at the course of events, it tends to include irrelevant or redundant frames (e.g., repeated instances of the man in the bottom example) because it lacks a notion of object importance.}
\label{fig:qualitative}
\end{figure*}

Overall, our summaries include more important people/objects with the same number of frames.  For example, for User 2 with selection criterion on $I(r) > 0.2$, our method finds 62\% of important objects in 27 frames, whereas the uniform keyframe and event-based adaptive keyframe sampling methods find 54\% and 46\% of important objects, respectively.  The lower absolute recall rate for all methods for User 4 is due to many small GT objects that appear together in the same frame (the user was cooking and baking).  On average, we find 9.13 events/video and 2.05 people/objects per event. 

While Fig.~\ref{fig:summarizationResults} captures the recall rate of the important objects, it does not measure the \emph{prominence} of the objects in the selected frames.  An informative summary should include not just any instance of the important object, but frames in which it is displayed prominently (i.e., large and centered).  To this end, in Fig.~\ref{fig:prominence}, we quantify the prominence of important objects in each method's summaries, in terms of the distance of the region's centroid to the frame center.  We see our method better isolates the prominent instances, thanks to its egocentric cues.  For example, in the top right example, the TV has high prominence in our summary and low prominence in the uniform keyframe sampling's summary.

\subsubsection{Summarization examples}

Fig.~\ref{fig:qualitative} shows example summaries from our method and the keyframe sampling baseline.  The colored blocks on ours indicate the discovered events.  We see that our summary not only has better recall of important objects, but it also selects views in which they are prominent in the frame.
 This helps more clearly reveal the story of the video.  For instance, for the top example, the story is: \emph{selecting an item at the supermarket $\rightarrow$ driving home $\rightarrow$ cooking $\rightarrow$ eating and watching TV}.  We provide additional summaries at the project webpage.

Fig.~\ref{fig:qualitative} (bottom) also depicts our method's failure modes.  Redundant frames of the same object can appear due to errors in event segmentation (see the man captured in Events 2 and 3) or the candidate important object clustering (the sink is captured twice in Event 10).  Adding features like GPS or depth might reduce such errors.

\begin{figure}[t]
\centering
\includegraphics[width=8cm]{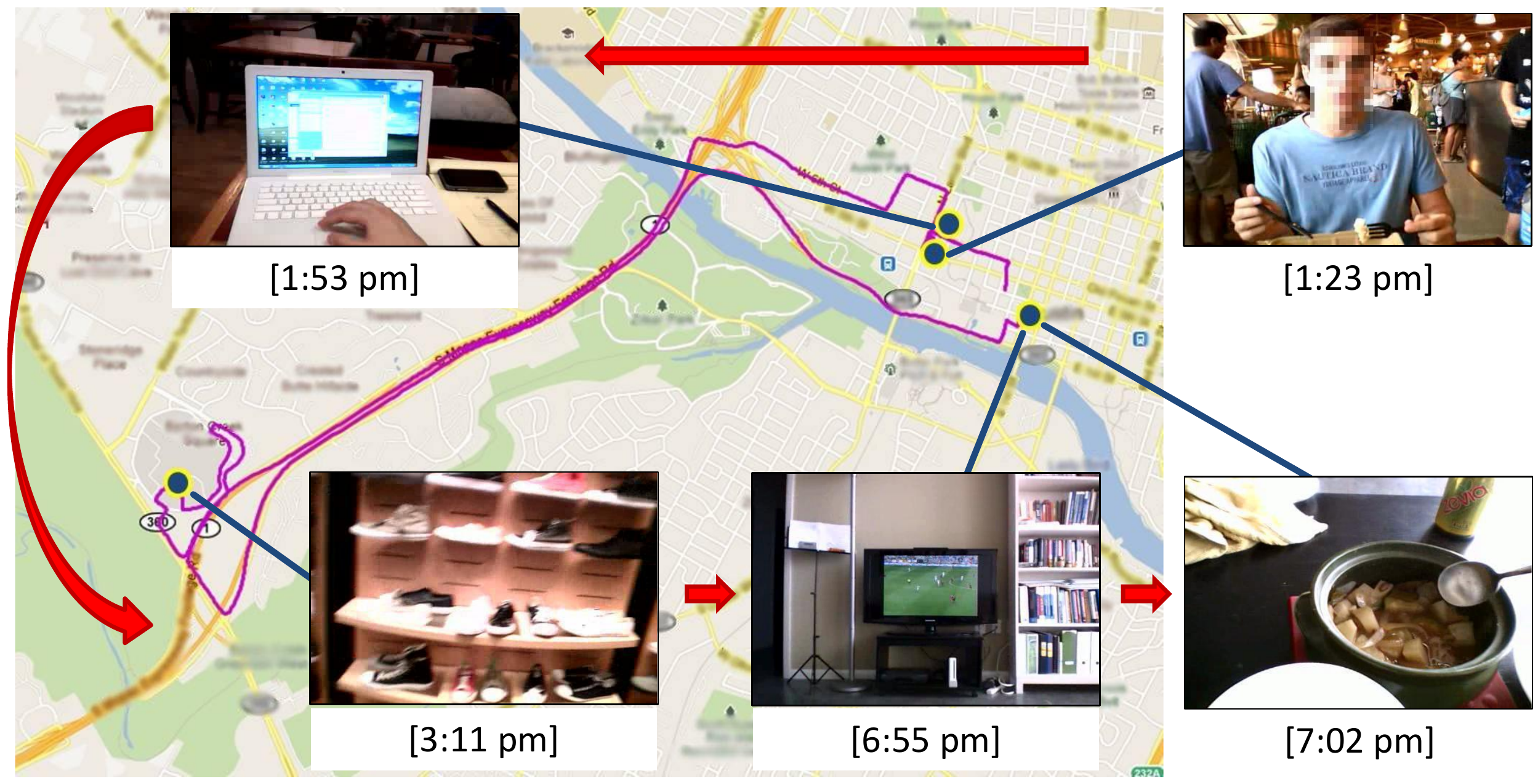}
\caption{An application of our approach that shows the GPS tracks of the camera wearer, the important people and objects that s/he interacted with, and their timeline.}
\label{fig:gps}
\end{figure}

Fig.~\ref{fig:gps} shows another example where we track the camera wearer's location with a GPS receiver, and display our method's keyframes on a map with the tracks (purple trajectory) and timeline.  This result suggests a novel multi-media application of our visual summarization algorithm that incorporates location, temporal, and visual data.

In all the results in this section, the two baselines perform fairly similarly to one another; compared to our method, they are prone to choosing unimportant or redundant frames that lack focus on those objects a human viewer has deemed important.  This supports our main hypothesis that the traditional low-level cues used in generic video summarization methods are insufficient to select keyframes that capture key objects in egocentric video.  Building on this finding, the user studies below analyze the impact that including important objects has on perceived summary quality.

\subsection{Budgeted frame selection accuracy}
\label{subsec:lengthBudgetAccuracy}

We next evaluate our approach for the scenario where we must handle requests such as, ``I would like to see a 10-frame summary of the original video''.

We compare our budgeted $k$-frame selection approach to four alternative methods: (1) the state-of-the-art video summarization method of~\cite{icme2009}, which selects keyframes that provide maximal content inclusion.  Briefly, it iteratively selects the frame that is on average most similar to all remaining frames without being redundant to the frames that have already been chosen.\footnote{This method summarizes a collection of videos, so we treat each event in our data as a different video.} (2) the keyframe selection approach of~\cite{Liu-2002}, which
optimizes an energy function that enforces adjacent frames to be maximally different. For fairest comparison, we use the same $\chi^2$-distance on color histograms used by our method to gauge visual dissimilarity. (3) a side-by-side implementation of our approach without event segmentation and region grouping (i.e., it selects $k$-frames from all frames of the video), and (4) uniform keyframe sampling.  The former two contrast our method with existing techniques that target the generic video summarization problem, highlighting the need to specialize to egocentric data as we propose.  The latter two isolate the impact of our importance predictions as well as our event segmentation and region grouping.

\subsubsection{Quantitative evaluation}

\begin{figure*}[t!]
\centering
\begin{tabular}{cccc}
\includegraphics[width=4.3cm]{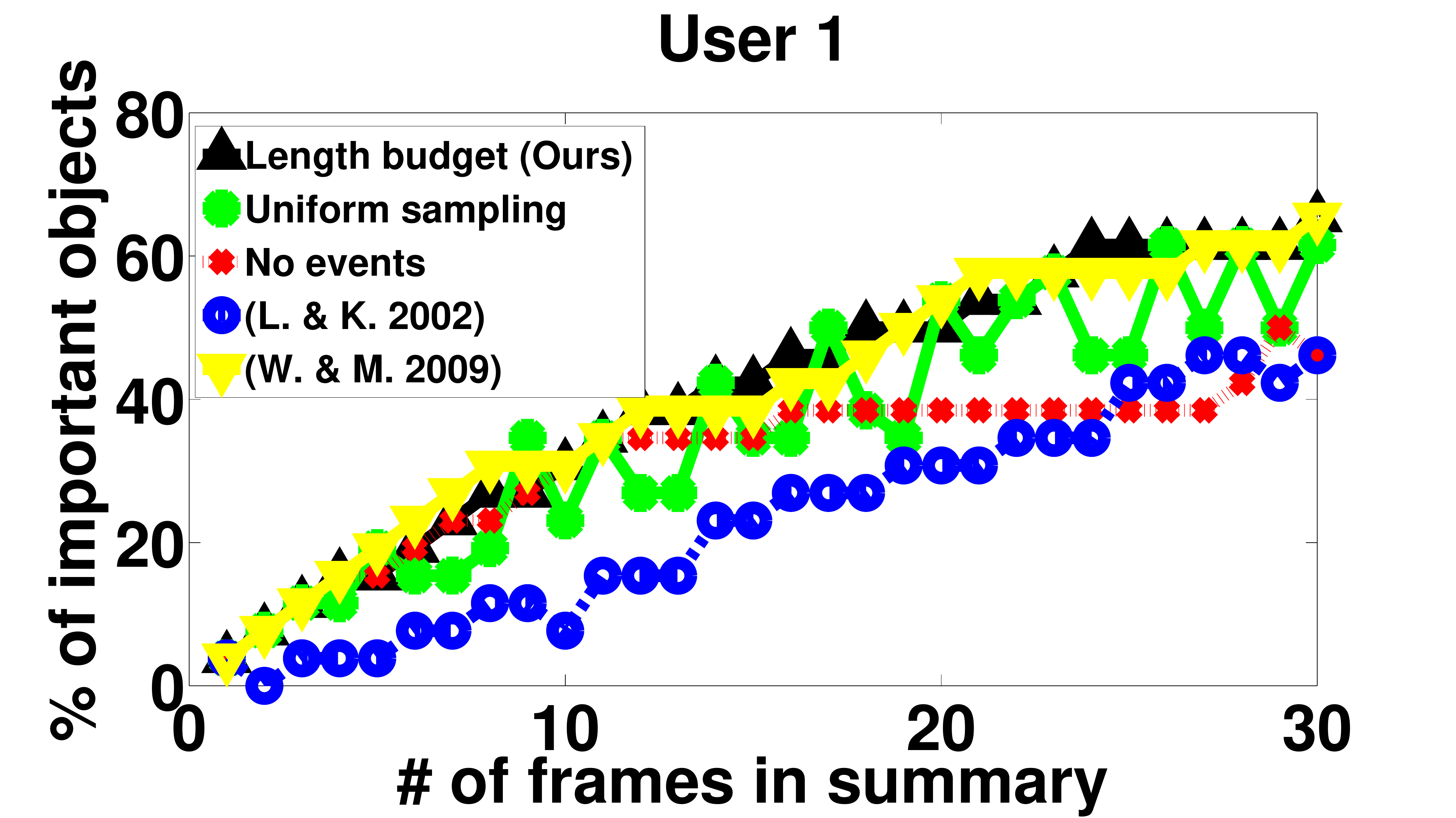}
\includegraphics[width=4.3cm]{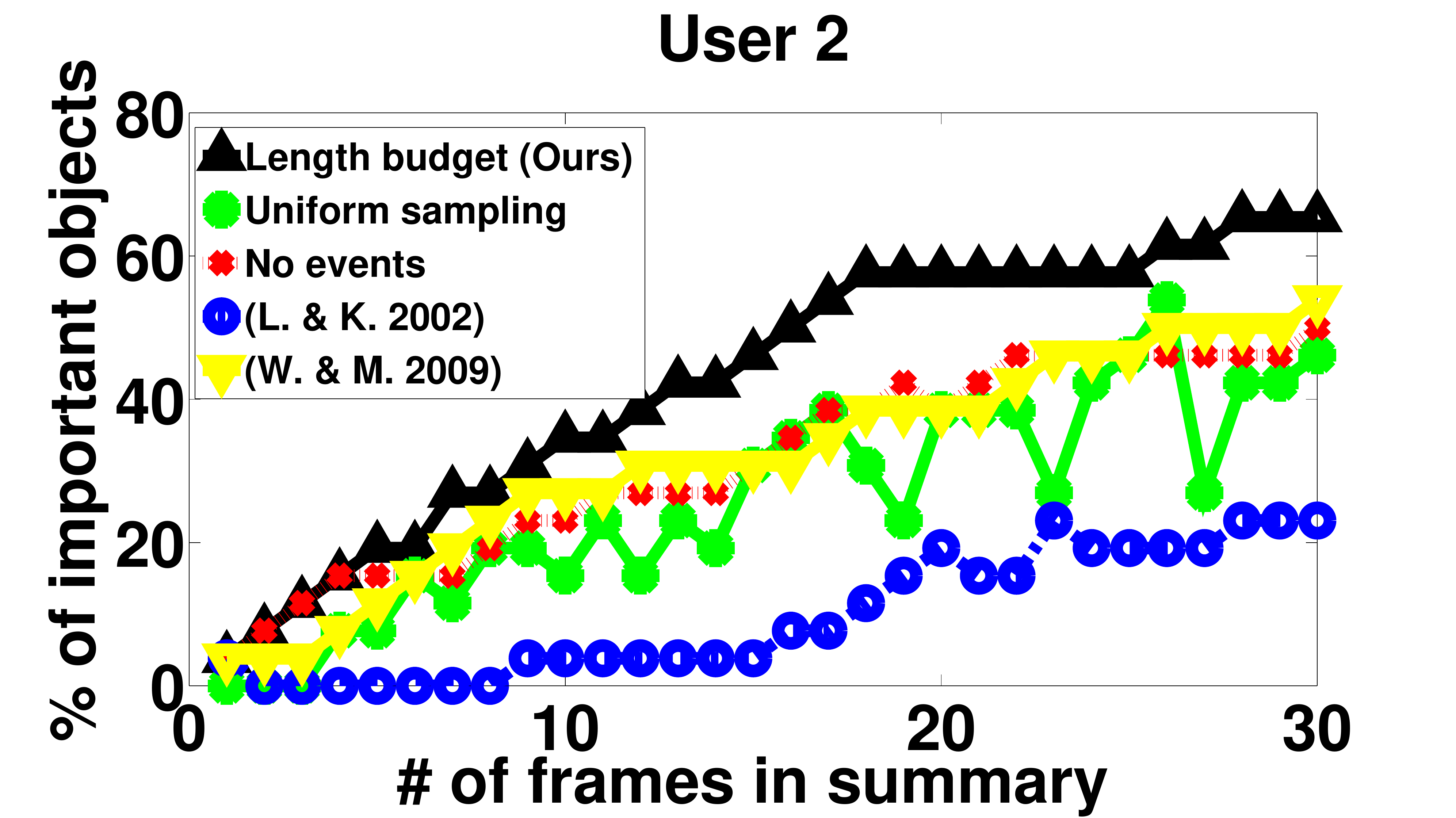}
\includegraphics[width=4.3cm]{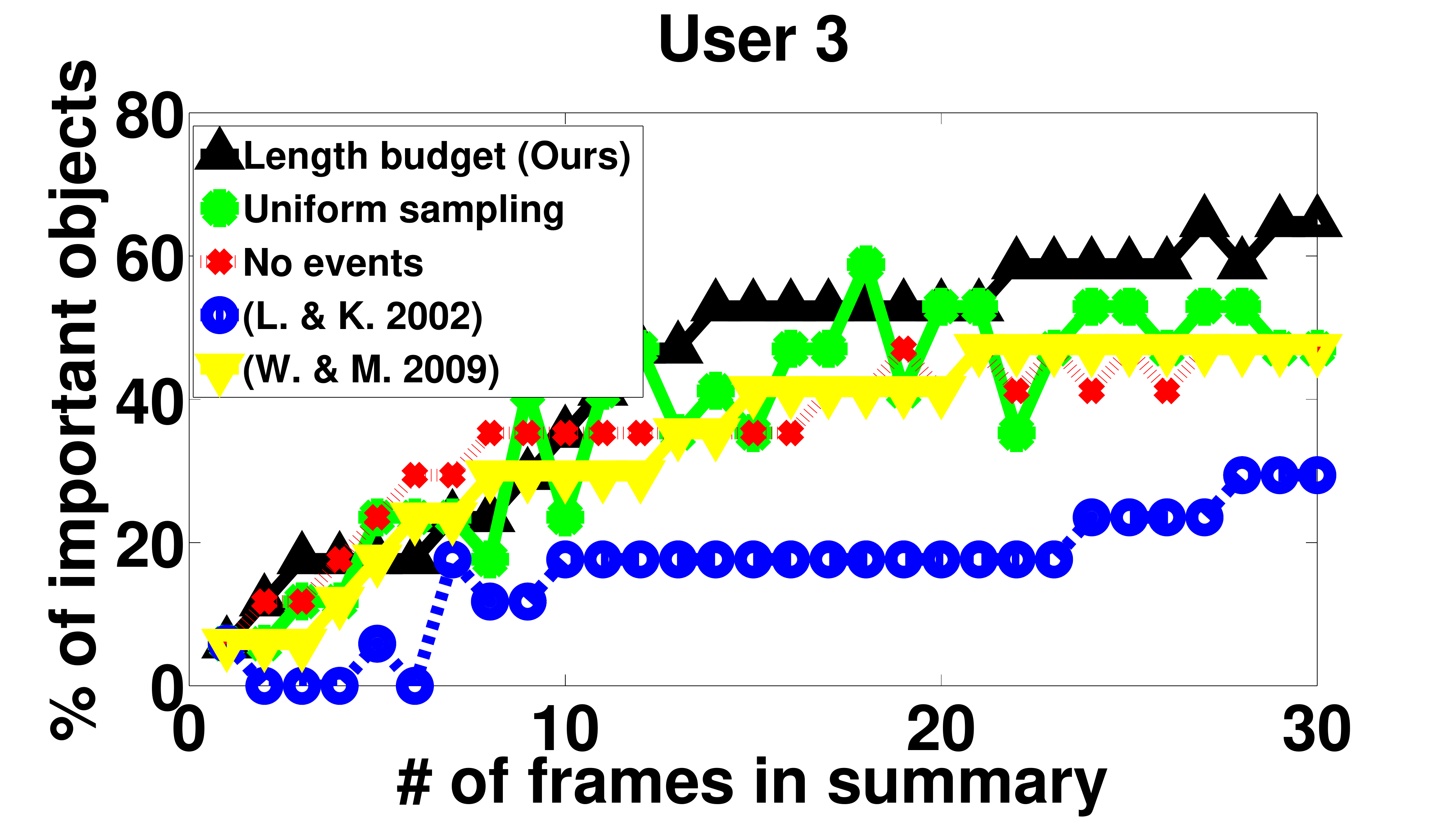}
\includegraphics[width=4.3cm]{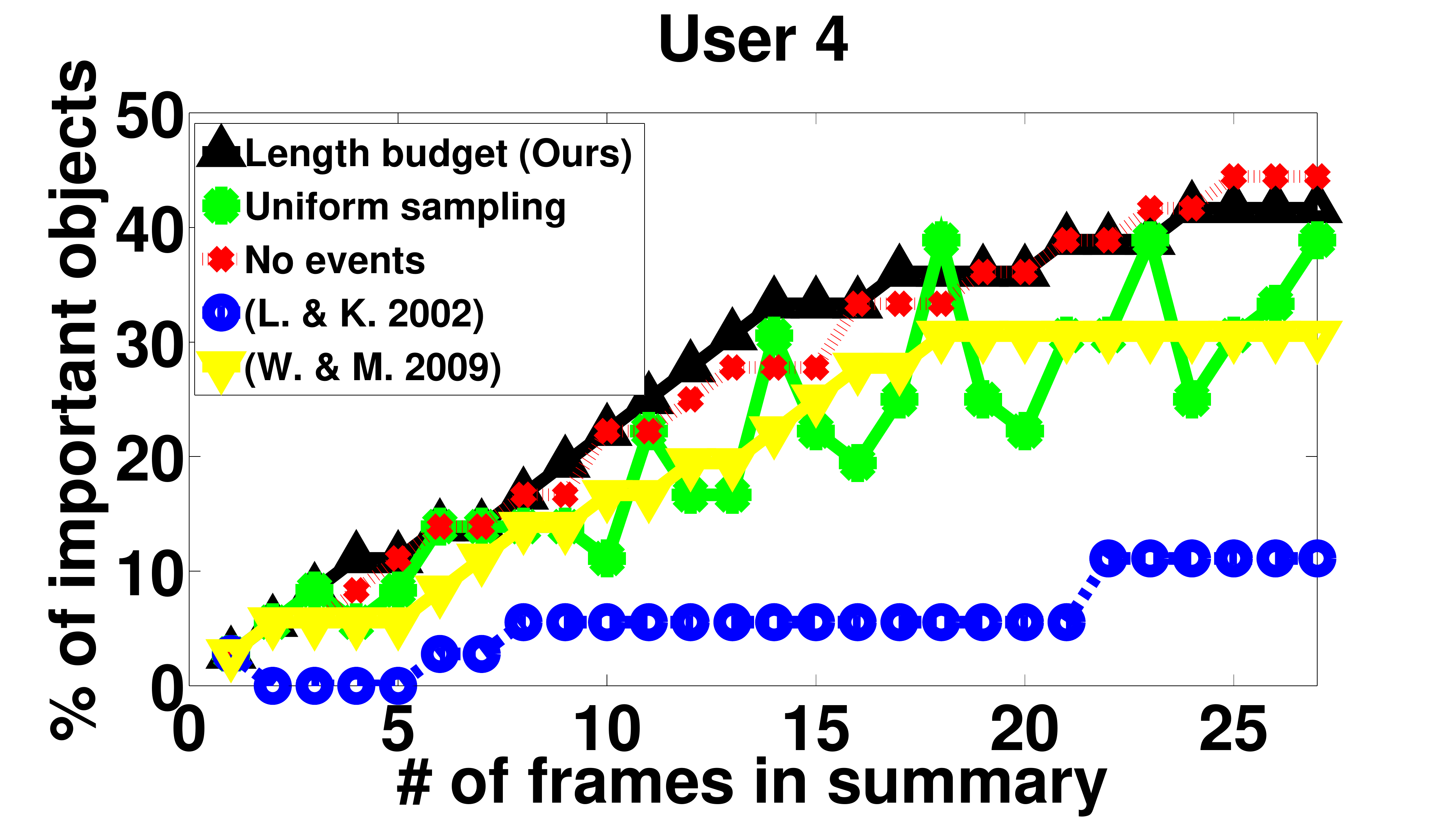}
\end{tabular}
\caption{Comparison to alternative $k$-frame summarization strategies.  Our budgeted frame selection approach produces more informative summaries with fewer frames.}
\label{fig:kframeResults} 
\end{figure*}

The plots in Fig.~\ref{fig:kframeResults} show the results.  We plot \emph{\% of important objects found} as a function of \emph{\# of frames in the summary}, in order to analyze both the recall rate of the important objects as well as the compactness of the summaries.  Each point on the curve shows the result for a different summary of the required length.  We score the objects found in the same way as in Section~\ref{subsubsec:quantEval1}.

Our model significantly outperforms the keyframe method~\cite{Liu-2002}, which confirms that modeling the importance of the object or person is critical to produce informative summaries for egocentric videos.  In fact, the existing method performs even worse than uniform sampling, due to its preference for frames that are maximally dissimilar to their surrounding selected frames.  As a result, it tends to select redundant frames containing the same visual elements in an alternating fashion.  Our summary does not have this issue since we represent each object in each event with a single region/frame through region clustering. 

Our model also outperforms the multi-document method~\cite{icme2009} on all but one user.  While this prior method successfully selects diverse content throughout the video, its reliance on low-level image cues leads to choosing some non-essential frames.

With very short summaries, uniform sampling performs similarly to ours; the selected keyframes are more spread out in time and have a high chance of including unique people/objects.  However, with longer summaries, our method always outperforms uniform sampling, since uniform sampling ignores object importance and tends to include frames repeating the same important object.

Our model also outperforms the baseline that selects $k$-frames from the entire video without event segmentation and region grouping (``No events'').  Since this method does not group instances of the same object together, it can select the same important object multiple times.

\subsubsection{Summarization examples}

\begin{figure*}[p!]
\centering
\includegraphics[width=17cm]{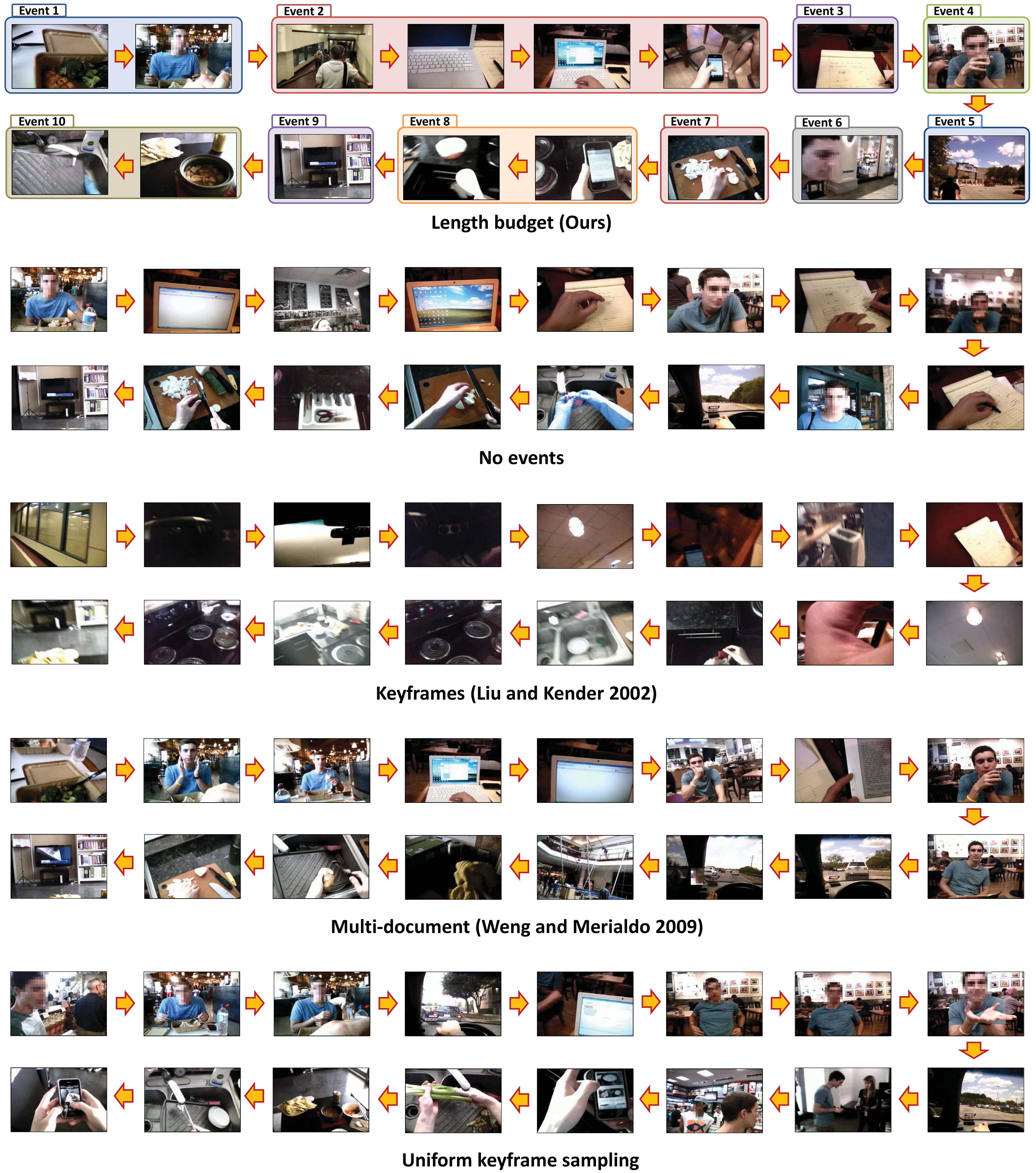}
\caption{Example summaries per method on UT Ego.  The ``No events'' baseline can include redundant frames because it lacks event segmentation and region clustering to group instances of the same object together (e.g., the yellow notepad and man).  Keyframe selection~\cite{Liu-2002} focuses on selecting adjacent frames that are maximally dissimilar, leading it to toggle between highly diverse frames, which need not capture important objects.  While the multi-document summarization objective~\cite{icme2009} overcomes this toggling effect, both it and uniform keyframe sampling tend to select redundant frames (e.g., see repeated instances of the man).  Overall, our summary best focuses on the important people and objects.  It selects informative frames that convey the chain of events through the objects and people that drive the first person interactions.}
\label{fig:kframe-qualitative}
\end{figure*}

Fig.~\ref{fig:kframe-qualitative} shows example summaries created by each method.  By focusing on the important people and objects, our method produces the best results.  

\subsection{User studies to evaluate summaries}
\label{subsec:userstudies}

We next perform user studies, since ultimately the impact of a summary depends on its value to a human viewer.  As subjects, we recruit both the camera wearers as well as 25 subjects uninvolved with the data collection or research in any way.  The camera wearers are a valuable resource to discern summary quality, since they alone fully experienced the original content.  Complementary to that, the uninvolved subjects are valuable to objectively gauge whether the overall events are understandable---without the implicit benefit of being able to ``fill in the gaps" with their own firsthand experience of the events being summarized.

\subsubsection{Evaluation by the camera wearers}

To quantify \emph{perceived} quality, we ask the camera wearers to compare our method's summaries to those generated by uniform keyframe sampling.  The camera wearers are good judges, since they know the full extent of their day that we are attempting to summarize.

We generate four pairs of summaries for each user, each of different length.  We ask the subjects to view our summary and the baseline's (in some random order unknown to the subject, and different for each pair), and answer two questions: (1) \emph{Which summary captures the important people/objects of your day better?} and (2) \emph{Which provides a better overall summary?}  The first specifically isolates how well each method finds important, prominent objects, and the second addresses the overall quality and story of the summary.  

\begin{table}[t!]
\hspace*{-0.05in}
\scriptsize
\centering
\begin{tabular}{|c|c|c|c|c|c|}
\hline
& Much better & Better & Similar & Worse & Much worse\\
\hline
Imp. captured & 31.25\% & 37.5\% & 18.75\% & 12.5\% & 0\% \\
\hline
Overall quality & 25\% & 43.75\% & 18.75\% & 12.5\% & 0\% \\
\hline
\end{tabular}
\caption{Camera wearer user study results.}
\label{table:user}
\end{table}

Table~\ref{table:user} shows the results, in terms of how often our summary is preferred.  In short, out of 16 total comparisons, our summaries were found to be better 68.75\% of the time.  We find our approach can fail to produce better summaries than uniform keyframe sampling if the user's day is very simple.  Specifically, User 3 was working on her laptop the entire day; first at home, then at class, then during lunch, and finally at the library.  For this video, uniform keyframe sampling was sufficient to produce a good summary.

\subsubsection{Evaluation by independent subjects}
\label{subsubsec:indsubject}

Next, to measure the quality of our summary on an absolute scale and to allow independent judges to evaluate a visual summary's informativeness, we ask each camera wearer to provide a ``ground-truth'' text summary of his/her day.  Specifically, we ask the users to provide full sentence descriptions that emphasize the key happenings (i.e., who s/he met, what s/he did, where, and when), and in sequential order as they happened that day. \yj{The resulting text summaries are 6-10 sentences long.  Here is an example from User 2:}

\emph{``My boyfriend and I drove to a farmers market in the early afternoon, where we sampled some food. Then (also in the early afternoon) we drove to a pizza place, where we stayed for a while, talked, had pizza, drank beer, and watched TV. After that, in the afternoon, we walked to a frozen yogurt place and split a cup of frozen yogurt, with brief looks at an animation that was playing. Then we walked around for a while, and drove home in the early evening. At home, we played with Legos for a while, in the living room. Then we watched some videos on YouTube. After that we played with Legos some more, and I washed some dishes in the kitchen, in the evening.''}  See the supplementary file for the remaining text summaries.

We then ask 25 subjects using Mechanical Turk to compare our summary and the baselines' (without knowing which method generated the summary) to the text summary provided by the camera wearer of the corresponding video, and answer: \emph{How well does the visual summary follow the story of the text summary?}  On a scale of 1 to 5 (1 being ``very well'' and 5 being ``very poorly''), over all 16 summaries, ours scored $2.61$ ($\pm0.97$).  The prior methods~\cite{Liu-2002},~\cite{icme2009}, and uniform sampling scored only $3.43$ ($\pm1.05$), $3.28$ ($\pm1.10$), $2.94$ ($\pm1.09$), respectively.  In general, the judges found the longer summaries to better align with the corresponding text summary than the shorter summaries.  On some videos, our shorter summaries failed to capture all of the details in the text summary, resulting in poor scores.

\begin{table}[t!]
\hspace*{-0.25in}
\scriptsize
\centering
\begin{tabular}{|c|c|c|c|c|c|}
\hline
& Much better & Better & Similar & Worse & Much worse\\
\hline
keyframes~\cite{Liu-2002} & 16.43\% & 45.45\% & 13.99\% & 18.88\% & 5.25\% \\
\hline
multi-document~\cite{icme2009} & 21.08\% & 36.14\% & 17.47\% & 17.47\% & 7.84\% \\
\hline
uniform sampling & 10.22\% & 37.63\% & 14.52\% & 29.03\% & 8.60\% \\
\hline
\end{tabular}
\caption{Mechanical Turk user study results on UT Ego.}
\label{table:user2}
\end{table}

\begin{table}[t!]
\hspace*{-0.25in}
\scriptsize
\centering
\begin{tabular}{|c|c|c|c|c|c|}
\hline
& Much better & Better & Similar & Worse & Much worse\\
\hline
keyframes~\cite{Liu-2002} & 26.80\% & 41.24\% & 17.01\% & 11.34\% & 3.61\% \\
\hline
multi-document~\cite{icme2009} & 13.90\% & 28.88\% & 25.67\% & 26.20\% & 5.35\% \\
\hline
uniform sampling & 11.95\% & 33.96\% & 20.75\% & 25.79\% & 7.55\% \\
\hline
\end{tabular}
\caption{Mechanical Turk user study results on ADL.}
\label{table:adl}
\end{table}

While the result above gauges quality on an absolute scale, we also ran a comparative test.  Here, we ask the subjects to compare our summary and each baseline's (in random order) to the text summary, and answer: \emph{Which visual summary more closely follows the story of the text summary?}  Table~\ref{table:user2} shows the accumulated responses from all 25 subjects.  Out of 16 total comparisons to each baseline, our summaries were found to be better 48-62\% of the time, and only worse 24-38\% of the time.

\subsection{Experiments on ADL}
\label{subsec:adl}

\yj{Finally, we perform experiments on ADL, an interesting and complimentary dataset to UT Ego that contains egocentric videos of people performing daily activities in their home (e.g., washing dishes, brushing teeth, etc.). It contains 20 videos, each roughly 30 minutes in length.}

\begin{figure*}[t!]
\centering
\includegraphics[width=17cm]{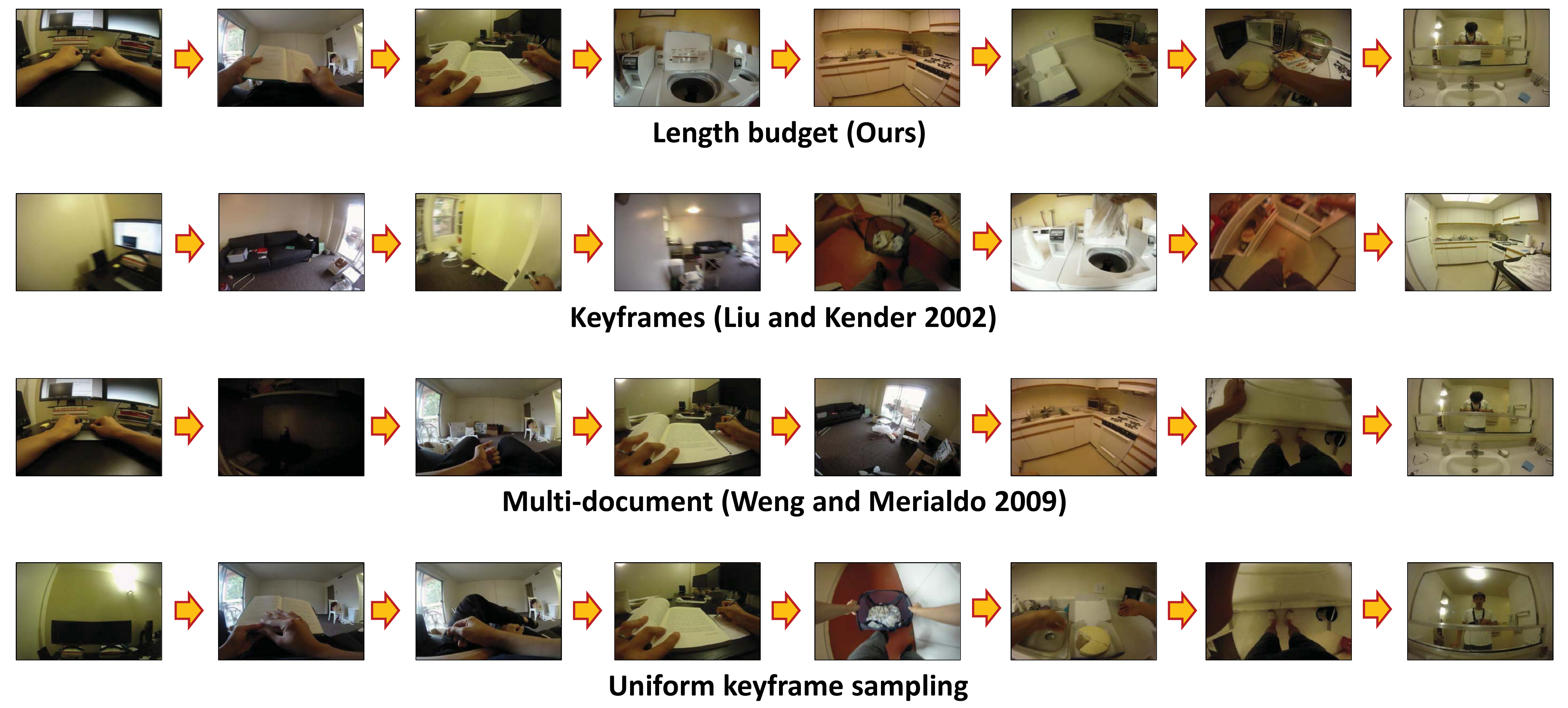}
\caption{Example summaries per method on ADL.  Keyframe selection~\cite{Liu-2002} focuses on selecting adjacent frames that are maximally dissimilar, leading it to toggle between highly diverse frames, which need not capture important objects.  While the multi-document summarization objective~\cite{icme2009} overcomes this toggling effect, both it and uniform keyframe sampling can select irrelevant frames (e.g., see 2nd and 7th columns).  Overall, our summary selects informative frames that best focus on the important objects that drive the first person interactions.}
\label{fig:adl-qualitative}
\end{figure*}

\yj{Since this data lacks ground-truth important object annotations, we use it only to evaluate our summaries.  We take the importance predictor from UT Ego (trained on all four videos), and use it to predict region importance on the ADL videos.  We use our budgeted frame selection approach and set the summary frame-length to $k=8$ (an arbitrary but reasonable number given the short length of ADL videos).  For each video, we ask an independent subject to watch the video and provide a text summary that emphasizes the key happenings, in the same manner as described in Sec.\ref{subsubsec:indsubject}.  The resulting summaries tend to focus on specific actions and are more descriptive than those provided by the camera wearers on UT Ego.  We suspect this is due to the relatively short length of each video ($\sim$3sll0 minutes).  Here is an example summary:}

\emph{``A guy brought his laundry basket to the laundry room to do laundry. He poured in the liquid detergent and did his laundry. He then went back home and started to play a video game on TV. The guy went into his room and turned on his laptop computer and looked at a picture of a monkey. The guy went into the bathroom to wash his face and brush his teeth. The guy is now in his kitchen and poured some juice to drink. He's looking at a list and checking off his list. The guy is making tea. The guy went into the bathroom to comb his hair. The guy cleaned his kitchen floor with a broom. He then went into his bedroom and put on his shoes.''}  See supplementary file for all text summaries.

\yj{We then ask 10 Mechanical Turk subjects per video to compare our summaries to those of uniform sampling, keyframes~\cite{Liu-2002}, and multi-document~\cite{icme2009}, and ask the same set of questions as in Sec.~\ref{subsubsec:indsubject}.  Table~\ref{table:adl} shows the results.  Out of 20 total pairwise comparisons to each baseline, our summaries were found to be better 42-68\% of the time, and worse 15-33\% of the time.  In terms of how each method's summary compares to the text summary, ours,~\cite{Liu-2002},~\cite{icme2009}, and uniform sampling scored $2.71$ ($\pm1.02$), $3.58$ ($\pm0.97$), $2.99$ ($\pm1.15$), $2.89$ ($\pm1.06$), respectively (recall that lower numbers are better: 1 being ``very well'' and 5 being ``very poorly'').  We show clear improvement over keyframes~\cite{Liu-2002}, which tends to simply oscillate between bright/dark frames.  Our improvements over uniform sampling and multi-document~\cite{icme2009} are less compared to those on UT Ego.  This is likely due to the ADL videos being shorter in length and more structured; in ADL, the camera wearers are given a list of actions they should perform, whereas UT Ego is completely unscripted.  Under these conditions, summarization algorithms that aim to select frames that are spread-out over time are likely to select meaningful frames.  Still, by focusing on the important objects, our approach produces the best summaries.  Fig.~\ref{fig:adl-qualitative} shows example summaries created by each method.  Our method selects the most informative frames.}

Overall, the results are a promising indication that discovering important people and objects leads to higher quality summaries for egocentric video.  Not only do we better recount those objects that human viewers deem important in the context of the surrounding activity, but we also generate summaries that human viewers prefer to multiple existing summarization approaches.

\section{Conclusion and Future Work}

We introduced an approach to summarize egocentric video using novel egocentric cues to predict important regions.  We presented two ways to adjust summary compactness: given either an importance selection criterion or a length budget.  For the latter, we developed an efficient optimization strategy to recover the best $k$-frame summary.  To our knowledge, ours is the first work to summarize videos from wearable cameras \yj{by discovering objects that may be important to the camera wearer}.  Existing summarization techniques rely on static cameras or low-level visual similarity, and so they fail to account for the key objects that drive first person interactions.  Through extensive experiments, we showed that our approach produces significantly more informative summaries than prior methods.

Future work can expand this idea in several interesting directions.  We assumed that the importance cues can be learned and shared across users, and our experiments confirmed that it is feasible.  However, there are also subjective elements; e.g., depending on the user, a person that he has significant interactions with may or may not be considered important.  To overcome the subjectivity, one could learn a wearer-specific model that uses input from the wearer for training to complement our wearer-independent model.

Secondly, event segmentation remains a challenge for egocentric data.  With the frequent head and body motion inherent to wearable video, grouping frames according to low-level scene statistics is imperfect.  In our system, this can sometimes lead to redundant keyframes showing the same object.  One way to mitigate this issue is to use a GPS receiver and generate event clusters using both location information and scene appearance.  This could provide better separation of events, especially when the scene appearance between two neighboring events is similar.  More broadly, more robust detection of event boundaries is needed.

Finally, while our interest lies in the computer vision challenges, other sensing modalities naturally can play a role in egocentric summarization.  For example, audio cues could signal person importance based on their speech near the camera, while ambient noise may be indicative of the scene type.  Other sensors like an accelerometer can reveal the user's gestures and activity, while GPS coordinates could give real-world location context relevant to which objects are likely important (e.g., a plate in a restaurant, vs. an athlete in a stadium).

\bibliographystyle{IEEEtran}
\bibliography{strings,refs}

\begin{thebibliography}{10}
\providecommand{\url}[1]{#1}
\csname url@samestyle\endcsname
\providecommand{\newblock}{\relax}
\providecommand{\bibinfo}[2]{#2}
\providecommand{\BIBentrySTDinterwordspacing}{\spaceskip=0pt\relax}
\providecommand{\BIBentryALTinterwordstretchfactor}{4}
\providecommand{\BIBentryALTinterwordspacing}{\spaceskip=\fontdimen2\font plus
\BIBentryALTinterwordstretchfactor\fontdimen3\font minus
  \fontdimen4\font\relax}
\providecommand{\BIBforeignlanguage}[2]{{%
\expandafter\ifx\csname l@#1\endcsname\relax
\typeout{** WARNING: IEEEtran.bst: No hyphenation pattern has been}%
\typeout{** loaded for the language `#1'. Using the pattern for}%
\typeout{** the default language instead.}%
\else
\language=\csname l@#1\endcsname
\fi
#2}}
\providecommand{\BIBdecl}{\relax}
\BIBdecl

\bibitem{wolf-1996}
W.~Wolf, ``Keyframe {S}election by {M}otion {A}nalysis,'' in \emph{ICASSP},
  1996.

\bibitem{zhang-pr1997}
H.~J. Zhang, J.~Wu, D.~Zhong, and S.~Smoliar, ``An {I}ntegrated {S}ystem for
  {C}ontent-{B}ased {V}ideo {R}etrieval and {B}rowsing,'' in \emph{Pattern
  Recognition}, 1997.

\bibitem{goldman-2006}
D.~Goldman, B.~Curless, D.~Salesin, and S.~Seitz, ``Schematic {S}toryboarding
  for {V}ideo {V}isualization and {E}diting,'' in \emph{SIGGRAPH}, 2006.

\bibitem{Liu-2002}
T.~Liu and J.~R. Kender, ``Optimization {A}lgorithms for the {S}election of
  {K}ey {F}rame {S}equences of {V}ariable {L}ength,'' in \emph{ECCV}, 2002.

\bibitem{aner-2002}
A.~Aner and J.~R. Kender, ``Video {S}ummaries through {M}osaic-{B}ased {S}hot
  and {S}cene {C}lustering,'' in \emph{ECCV}, 2002.

\bibitem{caspi-2006}
Y.~Caspi, A.~Axelrod, Y.~Matsushita, and A.~Gamliel, ``Dynamic {S}tills and
  {C}lip {T}railer,'' in \emph{The Visual Computer}, 2006.

\bibitem{ravacha-2006}
A.~Rav-Acha, Y.~Pritch, and S.~Peleg, ``Making a {L}ong {V}ideo {S}hort,'' in
  \emph{CVPR}, 2006.

\bibitem{pritch-2007}
Y.~Pritch, A.~Rav-Acha, A.~Gutman, and S.~Peleg, ``Webcam {S}ynopsis: {P}eeking
  {A}round the {W}orld,'' in \emph{ICCV}, 2007.

\bibitem{hodges-memory2011}
S.~Hodges, E.~Berry, and K.~Wood, ``Sensecam: {A} {W}earable {C}amera which
  {S}timulates and {R}ehabilitates {A}utobiographical {M}emory,''
  \emph{Memory}, 2011.

\bibitem{lee-2007}
M.~Lee and A.~Dey, ``Providing good memory cues for people with episodic memory
  impairment,'' in \emph{ACM SIGACCESS Conference on Computers and
  Accessibility}, 2007.

\bibitem{egocentric}
Y.~J. Lee, J.~Ghosh, and K.~Grauman, ``Discovering {I}mportant {P}eople and
  {O}bjects for {E}gocentric {V}ideo {S}ummarization,'' in \emph{CVPR}, 2012.

\bibitem{deva-egocentric-cvpr2012}
H.~Pirsiavash and D.~Ramanan, ``Detecting activities of daily living in
  first-person camera views,'' in \emph{CVPR}, 2012.

\bibitem{icme2009}
F.~Weng and B.~Merialdo, ``{Multi-Document Video Summarization},'' in
  \emph{ICME}, 2009.

\bibitem{simakov-cvpr2008}
D.~Simakov, Y.~Caspi, E.~Shechtman, and M.~Irani, ``Summarizing {V}isual {D}ata
  using {B}idirectional {S}imilarity,'' in \emph{CVPR}, 2008.

\bibitem{liu-pami2009}
D.~Liu, G.~Hua, and T.~Chen, ``A {H}ierarchical {V}isual {M}odel for {V}ideo
  {O}bject {S}ummarization,'' in \emph{TPAMI}, 2009.

\bibitem{itti-1998}
L.~Itti, C.~Koch, and E.~Niebur, ``A {M}odel of {S}aliency-based {V}isual
  {A}ttention for {R}apid {S}cene {A}nalysis,'' \emph{TPAMI}, vol.~20, no.~11,
  November 1998.

\bibitem{gao-nips2007}
D.~Gao, V.~Mahadevan, and N.~Vasconcelos, ``The {D}iscriminant
  {C}enter-{S}urround {H}ypothesis for {B}ottom-{U}p {S}aliency,'' in
  \emph{NIPS}, 2007.

\bibitem{liu-cvpr07}
T.~Liu, J.~Sun, N.~Zheng, X.~Tang, and H.~Shum, ``{L}earning to {D}etect a
  {S}alient {O}bject,'' in \emph{CVPR}, 2007.

\bibitem{objectness}
B.~Alexe, T.~Deselaers, and V.~Ferrari, ``What is an {O}bject?'' in
  \emph{CVPR}, 2010.

\bibitem{carreira-mincut}
J.~Carreira and C.~Sminchisescu, ``{C}onstrained {P}arametric {M}in-{C}uts for
  {A}utomatic {O}bject {S}egmentation,'' in \emph{CVPR}, 2010.

\bibitem{endres-eccv2010}
I.~Endres and D.~Hoiem, ``Category {I}ndependent {O}bject {P}roposals,'' in
  \emph{ECCV}, 2010.

\bibitem{keysegments}
Y.~J. Lee, J.~Kim, and K.~Grauman, ``Key-{S}egments for {V}ideo {O}bject
  {S}egmentation,'' in \emph{ICCV}, 2011.

\bibitem{spain-eccv2008}
M.~Spain and P.~Perona, ``Some {O}bjects are {M}ore {E}qual than {O}thers:
  {M}easuring and {P}redicting {I}mportance,'' in \emph{ECCV}, 2008.

\bibitem{hwang-bmcv2010}
S.~J. Hwang and K.~Grauman, ``Accounting for the {R}elative {I}mportance of
  {O}bjects in {I}mage {R}etrieval,'' in \emph{BMVC}, 2010.

\bibitem{clarkson-icassp1999}
B.~Clarkson and A.~Pentland, ``Unsupervised {C}lustering of {A}mbulatory
  {A}udio and {V}ideo,'' in \emph{ICASSP}, 1999.

\bibitem{starner-pami1998}
T.~Starner, J.~Weaver, and A.~Pentland, ``Real-time american sign language
  recognition using desk and wearable computer based video,'' \emph{PAMI},
  vol.~20, no.~12, pp. 1371--1375, 1998.

\bibitem{starner-iswc1998}
T.~Starner, B.~Schiele, and A.~Pentland, ``Visual contextual awareness in
  wearable computing,'' in \emph{ISWC}, 1998.

\bibitem{mann-1998}
S.~Mann, ``{Wearcam (the wearable camera): Personal imaging systems for long
  term use in wearable tetherless computer mediated reality and personal
  photo/videographic memory prosthesis},'' in \emph{Wearable Computers}, 1998.

\bibitem{healey-1998}
J.~Healey and R.~Picard, ``{Startlecam: A Cybernetic Wearable Camera},'' in
  \emph{Wearable Computers}, 1998.

\bibitem{hodges-2006}
S.~Hodges, L.~Williams, E.~Berry, S.~Izadi, J.~Srinivasan, A.~Butler, G.~Smyth,
  N.~Kapur, and K.~Wood, ``{SenseCam: A Retrospective Memory Aid},'' in
  \emph{UBICOMP}, 2006.

\bibitem{ng-icme02}
H.~W. Ng, Y.~Sawahata, and K.~Aizawa, ``Summarizing wearable videos using
  support vector machine,'' in \emph{ICME}, 2002.

\bibitem{Lin06a.:structuring}
W.~Lin and A.~Hauptmann, ``Structuring continuous video recordings of everyday
  life using time-constrained clustering,'' in \emph{IS\&T/SPIE Symposium on
  Electronic Imaging}, 2006.

\bibitem{doherty-2008}
A.~Doherty and A.~Smeaton, ``Combining face detection and novelty to identify
  important events in a visual lifelog,'' in \emph{International Conference on
  Computer and Information Technology Workshops}, 2008.

\bibitem{spriggs-wkshp}
E.~Spriggs, F.~D. la~Torre, and M.~Hebert, ``Temporal segmentation and activity
  classification from first-person sensing,'' in \emph{CVPR Workshop on
  Egocentric Vision}, 2009.

\bibitem{fathi-iccv2011}
A.~Fathi, A.~Farhadi, and J.~Rehg, ``Understanding {E}gocentric {A}ctivities,''
  in \emph{ICCV}, 2011.

\bibitem{ryoo-cvpr2013}
M.~S. Ryoo and L.~Matthies, ``{First-Person Activity Recognition: What Are They
  Doing to Me?}'' in \emph{CVPR}, 2013.

\bibitem{ren-cvpr2010}
X.~Ren and C.~Gu, ``Figure-{G}round {S}egmentation {I}mproves {H}andled
  {O}bject {R}ecognition in {E}gocentric {V}ideo,'' in \emph{CVPR}, 2010.

\bibitem{aghazadeh-cvpr2011}
O.~Aghazadeh, J.~Sullivan, and S.~Carlsson, ``Novelty {D}etection from an
  {E}gocentric {P}erspective,'' in \emph{CVPR}, 2011.

\bibitem{li-cvpr2013}
C.~Li and K.~M. Kitani, ``{Pixel-level Hand Detection for Ego-centric
  Videos},'' in \emph{CVPR}, 2013.

\bibitem{li-iccv2013}
Y.~Li, A.~Fathi, and J.~M. Rehg, ``{Learning to Predict Gaze in Egocentric
  Video},'' in \emph{ICCV}, 2013.

\bibitem{fathi-cvpr2012}
A.~Fathi, J.~K. Hodgins, and J.~M. Rehg, ``{Social Interactions: A First-Person
  Perspective},'' in \emph{CVPR}, 2012.

\bibitem{schiele-ubicomp08}
T.~Huynh, M.~Fritz, and B.~Schiele, ``Discovery of {A}ctivity {P}atterns using
  {T}opic {M}odels,'' in \emph{UBICOMP}, 2008.

\bibitem{jojic-2010}
N.~Jojic, A.~Perina, and V.~Murino, ``Structural {E}pitome: {A} {W}ay to
  {S}ummarize {O}ne's {V}isual {E}xperience,'' in \emph{NIPS}, 2010.

\bibitem{kitani-2011}
K.~Kitani, T.~Okabe, Y.~Sato, and A.~Sugimoto, ``Fast {U}nsupervised
  {E}go-{A}ction {L}earning for {F}irst-{P}erson {S}ports {V}ideo,'' in
  \emph{CVPR}, 2011.

\bibitem{doherty-civr2008}
A.~Doherty, D.~Byrne, A.~Smeaton, G.~Jones, and M.~Hughes, ``Investigating
  {K}eyframe {S}election {M}ethods in the {N}ovel {D}omain of {P}assively
  {C}aptured {V}isual {L}ifelogs,'' in \emph{CIVR}, 2008.

\bibitem{lu-cvpr2013}
Z.~Lu and K.~Grauman, ``{Story-Driven Summarization for Egocentric Video},'' in
  \emph{CVPR}, 2013.

\bibitem{skindetection}
M.~Jones and J.~Rehg, ``Statistical {C}olor {M}odels with {A}pplication to
  {S}kin {D}etection,'' \emph{IJCV}, vol.~46, no.~1, 2002.

\bibitem{pedro-superpixels}
P.~Felzenszwalb and D.~Huttenlocher, ``Efficient {G}raph-{B}ased {I}mage
  {S}egmentation,'' \emph{IJCV}, vol.~59, no.~2, 2004.

\bibitem{turk-fg2004}
M.~Kolsch and M.~Turk, ``Robust {H}and {D}etection,'' in \emph{F{G}}, 2004.

\bibitem{lowe}
D.~Lowe, ``Distinctive {I}mage {F}eatures from {S}cale-{I}nvariant
  {K}eypoints,'' \emph{IJCV}, vol.~60, no.~2, 2004.

\bibitem{viola}
P.~Viola and M.~Jones, ``Rapid {O}bject {D}etection using a {B}oosted {C}ascade
  of {S}imple {F}eatures,'' in \emph{CVPR}, 2001.

\bibitem{perona-eccv1998}
P.~Perona and W.~Freeman, ``A {F}actorization {A}pproach to {G}rouping,'' in
  \emph{ECCV}, 1998.

\bibitem{bing}
M.-M. Cheng, Z.~Zhang, W.-Y. Lin, and P.~Torr, ``{BING: Binarized Normed
  Gradients for Objectness Estimation at 300fpsn},'' in \emph{CVPR}, 2014.

\bibitem{brox-pami2011}
T.~Brox and J.~Malik, ``Large {D}isplacement {O}ptical {F}low: {D}escriptor
  {M}atching in {V}ariational {M}otion {E}stimation,'' \emph{TPAMI}, vol.~33,
  no.~3, pp. 500--513, 2011.

\bibitem{walther-nn2006}
D.~Walther and C.~Koch, ``Modeling {A}ttention to {S}alient
  {P}roto-{O}bjects,'' \emph{Neural Networks}, vol.~19, pp. 1395--1407, 2006.

\bibitem{summary-survey-2008}
A.~Money and H.~Agius, ``Video summarisation: A conceptual framework and survey
  of the state of the art,'' \emph{Journal of Visual Communication and Image
  Representation}, vol.~19, no.~2, pp. 121--143, Feb 2008.

\end{thebibliography}

\end{document}